%% file: main.tex
\documentclass{article}

\PassOptionsToPackage{numbers, compress}{natbib}
\usepackage[preprint]{neurips_data_2024}

\usepackage[table]{xcolor}
\usepackage{colortbl}
\usepackage{xcolor}
\usepackage{pgf}
\usepackage{tabularx}
\usepackage{caption}

\usepackage{soul}  % Used for strikethrough

\usepackage{colortbl}
\definecolor{DnCBG}{rgb}{0.9, 0.9, 1.}  % blue 

\usepackage[utf8]{inputenc} % allow utf-8 input
\usepackage[T1]{fontenc}    % use 8-bit T1 fonts
\usepackage{hyperref}       % hyperlinks
\usepackage{url}            % simple URL typesetting
\usepackage{booktabs}       % professional-quality tables
\usepackage{amsfonts}       % blackboard math symbols
\usepackage{nicefrac}       % compact symbols for 1/2, etc.
\usepackage{microtype}      % microtypography
\usepackage{xcolor}         % colors

\usepackage{graphicx}

\usepackage{amsmath}
\usepackage{amssymb}

\usepackage{colortbl}
\definecolor{ourbg}{rgb}{0.9, 0.9, 1.}  % blue 
\definecolor{darkgray}{rgb}{0.43,0.43,0.43}  % blue 
\definecolor{grayarea}{rgb}{0.95, 0.95, 0.96}
\definecolor{llm}{rgb}{0.97, 0.97, 0.93}
\definecolor{ouralg}{rgb}{0.8, 0.8, 0.9}

\usepackage{algorithm}
\usepackage{listings}
\usepackage{pifont}% http://ctan.org/pkg/pifont
\newcommand{\xmark}{\ding{55}}%
\usepackage{multirow}

\usepackage{bm}

\def\vx{{\bm{x}}}

\def\vz{{\bm{z}}}
\def\x{{$\times$}}

\title{Data curation via joint example selection \\ further accelerates multimodal learning} % contrastive

\author{%
  Talfan Evans$^*$ \quad Nikhil Parthasarathy$^*$ \quad Hamza Merzic \quad Olivier J. Hénaff$^*$ \\ % \thanks{Equal Contribution\vspace{-1em}}
  \vspace{-0.75em} \\
  Google DeepMind, London, UK\\
}

\begin{document}

\maketitle

\input{Sections/abstract}
\input{Sections/introduction}
\input{Sections/related_work}

\input{Sections/methods}
\input{Sections/experiments}
\input{Sections/discussion}

\bibliographystyle{plainnat}
\bibliography{main.bib}

\clearpage
\appendix 
\input{Sections/appendix}

\end{document}

%% file: Sections/abstract.tex
\begin{abstract}
   
Data curation is an essential component of large-scale pretraining. 
In this work, we demonstrate that jointly 
% prioritizing
selecting
\textit{batches} of data is more effective for learning than selecting examples independently. Multimodal contrastive objectives expose the dependencies between data and thus naturally yield criteria for measuring the \textit{joint learnability} of a batch. We derive a simple and tractable algorithm for selecting such batches, which significantly accelerate training beyond individually-prioritized data points. 
%As performance improves by selecting from large super-batches, 
As performance improves by selecting from larger super-batches, we also leverage recent advances in model approximation to reduce the associated computational overhead. %As a result, 
As a result, our approach---multimodal contrastive learning with joint example selection (JEST)---surpasses state-of-the-art models with up to 13$\times$ fewer iterations and 10$\times$ less computation. Essential to the performance of JEST is the ability to steer the data selection process towards the distribution of smaller, well-curated datasets via pretrained reference models, exposing the level of data curation as a new dimension for neural scaling laws. 
\end{abstract}
% pretraining
% Essential to the performance of JEST is the ability to steer the data selection process towards the distribution of smaller, well-curated datasets via pretrained reference models, exposing data curation as a new dimension for neural scaling laws. 
% \olivier{
% }

%% file: Sections/introduction.tex
\section{Introduction}

Data quality is an essential driver of performance for large-scale pretraining. 
Whether in language \citep{gunasekar2023textbooks}, vision \citep{evans2023bad}, or multimodal modeling \citep{abbas2023semdedup,hessel2021clipscore,mahmoud2023sieve}, training on well-curated datasets has
%yielded better results than significantly larger, uncurated ones, at a fraction of the cost. 
consistently demonstrated that strong performance can be achieved with significantly less data.
%consistently demonstrated that strong performance can be achieved with significantly less data.
%\cite{burns2023weak}. 
%While much progress has been made in developing heuristics for curating particular sources of data,
However, %data must also be available in sufficient quantity to train large models and
%size is also important for training large models. Unfortunately, 
current data pipelines rely heavily on manual curation, which is difficult and expensive to scale. In contrast, model-based data curation \citep{loshchilov2015online,mindermann2022prioritized}, which uses features of the model being trained to select high quality data, holds promise for improving the slow, power-law scaling of large-scale pretraining across modalities, both in theory \citep{sorscher2022beyond} and in practice \citep{evans2023bad}. 

% \talfan{Do we need to point out exactly where our method differs to coresets etc. here?} \olivier{I think that's better in the RW} 
Existing methods apply curation at the level of individual data points \cite{coleman2019selection, sachdeva2024train}. 
Yet the quality of a batch is also a function of its composition, in addition to the summed quality of its data points considered independently.
% Heuristics such as hard negative mining \cite{mishchuk2017working,robinson2020contrastive,simo2015discriminative,wu2017sampling}, redundancy reduction \cite{abbas2023semdedup,sorscher2022beyond}, and coresets \cite{campbell2018bayesian,coleman2019selection,har2004coresets} have all been found to improve the composition of batches and accelerate learning, \talfan{but use scoring heuristics that are decoupled from the training signal}.
In computer vision, hard negatives (i.e.\ clusters of points which lie close to one another but contain different labels) have been found to provide a more effective learning signal than trivially solvable ones \citep{bucher2016hard, harwood2017smart,mishchuk2017working,robinson2020contrastive,simo2015discriminative,wu2017sampling,xuan2020hard}. 
In this work we seek to generalize this notion by asking whether model-based data-selection criteria applied to batches of data can accelerate learning beyond what is possible by selecting examples independently. % \talfan{In doing so, we demonstrate that `foundation dataset' approaches, where data is filtered once are sampled from independently thereinafter, are outperformed by batch-wise selection which can only be performed online.} \olivier{This seem a bit nitty-gritty: easy-reference JEST can be performed online, yet performs batch-level selection.}
% \talfan{HNM requires positives }

% \olivier{Talfan, PTAL }
In multimodal learning, contrastive objectives directly expose the interactions between examples in a batch. We therefore derive a simple and tractable algorithm for % data curation, 
%for \textit{online data curation} based on 
joint example selection---JEST---which efficiently selects relevant `sub-batches' of data from much larger `super-batches' given their model-based scores. 
%When trained with such  batches 
%, we find 
%multimodal learning is
%to be 
%greatly accelerated, reaching similar performance to state-of-the-art models \cite{zhai2023sigmoid} with up to 13\x \ fewer training iterations. 
%When applied to \textit{offline data curation} (selecting easy batches according to a pretrained reference model), our algorithm also delivers significant gains \olivier{Be more specific here}.
%\talfan{Our method can be used for batch selection of both pre-cached `reference' scores \cite{hessel2021clipscore} and \textit{learnability} based scoring \cite{mindermann2022prioritized}, which also takes as input the current state of the learner model.}
%\talfan{Our method also improves the performance of CLIP-Scoring \cite{hessel2021clipscore} and shows that improved training efficiency can be obtained from pre-scoring instead of pre-filtering of data ahead of training.}
%\olivier{CLIP-scoring is an independent examples selection method, so JEST + CLIP-scoring seems mutually exclusive.}
%Our best method produces greatly accelerates multimodal learning, reaching similar performance to state-of-the-art models \cite{zhai2023sigmoid} with up to 13\x \ fewer training iterations. }
% \talfan{Our selection method significantly improves performance relative to CILP-scoring \cite{hessel2021clipscore}. However, it was recently demonstrated that pre-filtering datasets underperforms}
When scoring batches with a pretrained reference model (i.e. \textit{easy-reference}), JEST accelerates learning relative to uniform batch selection, significantly improving on independent example selection using the same reference (as in CLIPScore \cite{hessel2021clipscore}). 
% Data selection based on good image-text alignment as evaluated by a pre-trained reference model (CLIPScore \cite{hessel2021clipscore}), which we call \textit{easy-reference} scoring, is significantly improved when combined with our joint example selection method, producing a 5\x speed-up in training iterations over IID batch selection.
%However, it was recently demonstrated that \textit{easy-reference} scoring will constrain the performance of larger scale runs by limiting the available training pool to data that is solved by the pre-trained model. We therefore 
When scoring batches according to their \textit{learnability}, which also takes into account the online model loss \cite{mindermann2022prioritized}, JEST improves further, matching the performance of state-of-the-art models \cite{zhai2023sigmoid} with up to 13\x \ fewer training iterations.
% We also apply our joint example selection method to \textit{learnability} scoring \cite{mindermann2022prioritized}, which also takes into account the current state of the learner model, producing similar performance to state-of-the-art models \cite{zhai2023sigmoid} with up to 13\x \ fewer training iterations.

%\talfan{WDYT about adding a sentence here that logically dismisses easyref for large scale training?}
Discovering highly learnable batches requires sifting through much larger super-batches of raw data. 
% In our experiments, \textit{learnability} scoring produced superior performance to \textit{easy-reference} and continues to improve when sifting through much larger super-batches of data, but requires inference passes from the learner model. 
% therefore
We make learnability scoring of large batches tractable by leveraging recent advances in online model approximation,
which reduce computation while still providing useful predictions \citep{beyer2023flexivit,li2023scaling, zhang2020accelerating}. % dehghani2024patch
By training a single model at multiple resolutions in parallel, we efficiently apply the model for scoring large super-batches, find their most learnable sub-batch, and spend more valuable computation for learning on them. Thanks to savings in both learning and example scoring, we reduce the overhead of scoring from 133\% to 10\% additional FLOPs while maintaining significant gains in training efficiency. This approximate scoring framework---Flexi-JEST---produces state-of-the-art models with 11\x \ fewer iterations \textit{and} 10\x \ fewer FLOPs.

Finally, we find that central to the performance of our framework is the ability to steer the curation process towards the distribution of smaller, well-curated datasets. This occurs naturally with the model-based selection criteria we consider through the concept of a pretrained reference model, which prioritizes examples 
%from large-uncurated datasets 
that most resemble the data it was trained on. Crucially, we find this process to enable strong \textit{data quality bootstrapping}: a reference model trained on a small curated dataset can effectively guide the curation of a much larger dataset, allowing the training of a model which strongly surpasses the quality of the reference model on many downstream tasks. % \talfan{Can we make a statement here about how scalable bootstrapping is a unique property of learnability? I.e. communicate our intuition that as the learner becomes proficient beyond the reference model, it is able to steer learning towards unlearned data, whereas CLIPscoring will continue to emphasize the same data points?} \olivier{What would we be basing this off of?}

\begin{figure*}
    \centering
    \includegraphics[width=1.0\linewidth]{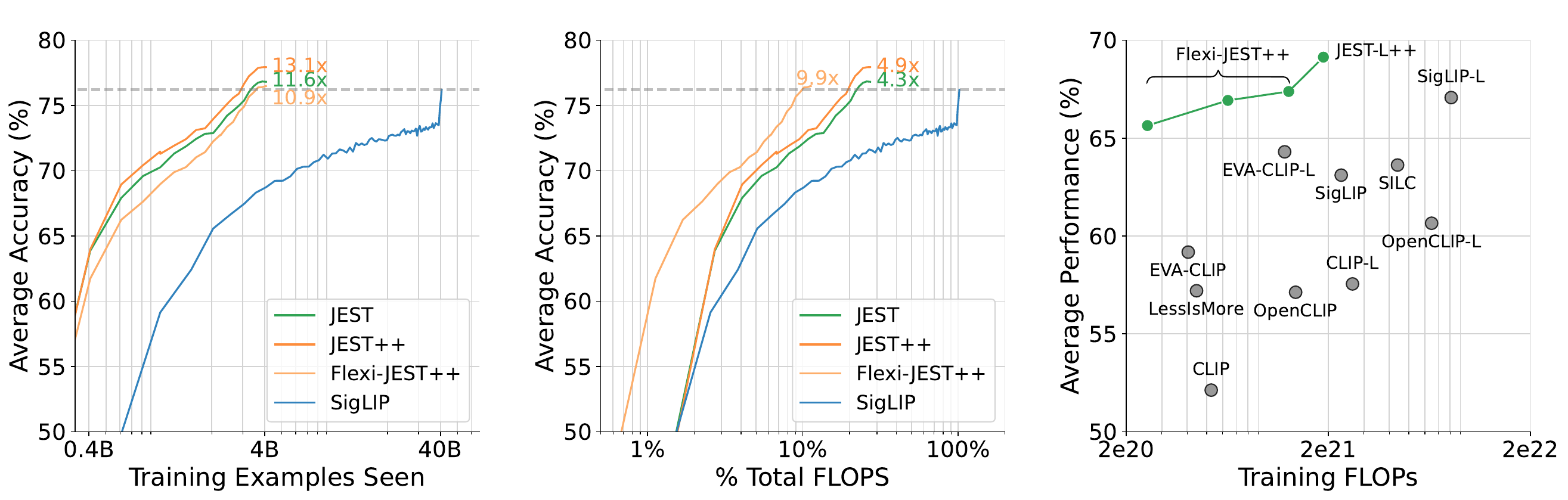}
    \vspace{-0.5em}
    \caption{
    \textbf{Joint Example Selection accelerates multimodal pretraining.} Our JEST/JEST++ methods bootstrap from small, strongly curated datasets (Webli-curated/Webli-curated++) to actively curate web-scale datasets. Flexi-JEST++ uses variable patch sizing to reduce the cost of curation. \textbf{Left}: Training with JEST matches the performance of the uniform 40B SigLIP baseline with up to 13$\times$ fewer iterations. \textbf{Middle}: Even when accounting for the cost of scoring, our best variant is almost 10$\times$ more FLOP efficient. \textbf{Right}: Comparison of JEST++/FlexiJEST++ (green) to prior methods (grey). Average accuracy is computed across 8 downstream tasks (left, middle; see Table \ref{tab:appendix_table_2}), or ImageNet and COCO (right).  \vspace{-1.em} % FLOP efficiency compard against external benchmarks.  COCO I2T, T2I, IN 0-Shot and 10-Shot right panel shows average accuracy over  
    }
    \label{fig:fig_speedups_teaser}
        \vspace{-1.em}
\end{figure*}
% Comparing the efficiency of three variants of our method against the 40B SigLIP model \cite{zhai2023sigmoid}. 
% (see Section \ref{sec:curation})
%  use reference models trained on different curated datasets, 
% is significantly more efficient than uniform sampling, and 
% (each Flexi-JEST update costs 1.1$\times$ the FLOPs of an equivalent IID update, see Appendix Section \ref{sec:flop_calcs}).

%% file: Sections/related_work.tex
\section{Related Work}

% \subsection{Offline curation of web-scale data}

\textbf{Offline curation: example-level data pruning.} Methods for collecting and filtering large-scale noisy image-text data initially focused on the quality of the textual captions \citep{byeon2022coyo,changpinyo2021conceptual,jia2021scaling}, and proximity to high-quality reference datasets \cite{fang2023data,gadre2023datacomp,xu2023demystifying}. Instead, model-based 
filtering approaches use pretrained models (such as CLIP \citep{radford2021learning} and BLIP \cite{li2022blip}) % as reference-free \talfan{reference-free?} 
as evaluation metrics for curating data via image-text alignment \citep{fang2023data, hessel2021clipscore, mahmoud2023sieve}. Critically, all of these methods are applied independently across examples, which fails to account for the relevance of dependencies across examples in a batch. % the usefulness of an example conditioned on what other examples are in the dataset. 
% \olivier{We seem to be missing some references here. Nikhil would you mind checking the Sieve paper and adding any references from their RW section? https://arxiv.org/pdf/2310.02110}
%\olivier{Still missing citations?}

\textbf{Offline curation: cluster-level data pruning.}
Other methods such as semantic redundancy reduction \citep{abbas2023semdedup, abbas2024effective,sorscher2022beyond} or core-set selection \cite{campbell2018bayesian,har2004coresets} have proposed to curate based on the marginal importance of data points given other data points in their vicinity. However these methods are based on a heuristic that is decoupled from the training objective% \talfan{Repeated, maybe remove from intro?}
. In contrast, our method enables joint-example selection that is specifically tailored to accelerating the contrastive pretraining objective function.

% \olivier{Suggest folding these references in the two paragraphs above. Can we just describe these as using "proximity to reference datasets"?} In addition to the literature on data curation, there have also been several studies that attempt to understand and design for characteristics of high-quality dataset, for example by studying the trade off between quality and quantity during data collection \citep{xu2023demystifying, fang2023data, goyal2024scaling}. Our work builds on these findings while also extending their scope in the context of dynamic curation in active learning frameworks.

\textbf{Online data curation with model-based scoring.} Pre-filtering using the curation procedures described above can lead to large increases in data quality. However, % \cut{in the large-data/few-epoch regime now popular in large-model training \citep{hoffmann2022training, kaplan2020scaling}, static curation can be nearly as expensive as pretraining itself. In addition,} 
fixed curation strategies do not take into account that the relevance of a training example can change over the course of learning, limiting their utility at scale \cite{goyal2024scaling}. These concerns are addressed by %model-based
online data curation methods \citep{evans2023bad,lin2024rho,loshchilov2015online,mindermann2022prioritized}, which identify high-quality examples \textit{not yet learned by the model}. Our work generalizes these by applying %the same
model-based criteria to batch-level (rather than example-level) losses, and selecting data accordingly. % , which we find to yield signficiant benefits.

\textbf{Hard negative mining.} A long literature has described the efficiency gains afforded by choosing the right set of negative examples in classical metric-learning \citep{bucher2016hard, harwood2017smart,mishchuk2017working,simo2015discriminative,wu2017sampling,xuan2020hard} as well as modern contrastive learning \cite{robinson2020contrastive, tian2021divide}. 
%We revisit these ideas by generalizing the notion of ``hard negatives'' to that of \textit{learnable negatives} which are hard for the learner but easy for a pretrained reference model.  
We generalize hard negative mining in two ways: 1)
we jointly mine for both positive and negative pairs by selecting entire batches and %(removing the need to condition on positive samples)
% we remove the need to condition on positive `anchor' samples by performing true joint selection  based on blocked Gibbs sampling and 
2) we explore prioritizing \textit{learnable} negatives, which are hard for the learner but easy for a pretrained model.

\textbf{Model approximation.} Several works have demonstrated that smaller models can be used as proxies for much larger models for data selection  \cite{coleman2019selection,evans2023bad,xie2023doremi}. However, several techniques have recently been developed that allow inference-time trade-offs between computation and performance, allowing smaller models to be ``embedded'' without the need for separate training. For Vision Transformers \cite{dosovitskiy2020image}, dropping patches \cite{li2023scaling} or layers \cite{zhang2020accelerating}, or reducing token resolution \citep{beyer2023flexivit} produce characteristic trade-offs \cite{li2024inverse}. Our work is the first to use these techniques in the context of online data selection.

%% file: Sections/methods.tex
\section{Methods}
\label{sec:methods}
\subsection{Model-based batch-selection criteria}
\label{sec:mps}
% \talfan{I think we just need to state the filtering fraction / super- / sub-batch definitions right away }
We refer to the model which we are interested in training as the \textit{learner}. Assuming we have a ``super-batch'' $\mathcal{D}$ (of size $B$) %  = |\mathcal{D}|
examples to learn from, we wish to extract a sub-batch $\mathcal{B} = \{ \vx_i, i \in [1, ..., b] \} \subset \mathcal{D}$ that is maximally relevant for learning. Prioritized sampling \cite{loshchilov2015online,schaul2015prioritized} performs this by scoring individual examples, then sampling in proportion to these scores. In this work we instead score entire sub-batches, and sample according to these batch-level scores. We consider model-based scoring functions, which use the losses from the learner model and/or pretrained \textit{reference} models.

\textbf{Hard learner.} An intuitive heuristic would be to prioritize batches $\mathcal{B}$ that have a high loss under the learner with parameters $\theta$: $s^\textrm{hard}( \mathcal{B} | \theta)  =  \ell( \mathcal{B} | \theta)$, which has the desirable property of discarding trivial data. This heuristic has been proven to work for small, clean datasets \cite{paul2021deep, sorscher2022beyond} but tends to do more harm than good for larger, less curated datasets \cite{evans2023bad} since it will also up-sample noisy examples.

\textbf{Easy reference.} In contrast, one could also choose to up-sample data that is ``easy'' (has low loss) for a pretrained \textit{reference} model with parameters $\theta^*$: $s^\textrm{easy}( \mathcal{B} | \theta^*)  =  - \ell( \mathcal{B} | \theta^*)$. This \textit{easy reference} heuristic has been used successfully in multimodal learning to identify high-quality examples \cite{hessel2021clipscore,schuhmann2022laion}, but does not reflect the current state of the learner and can therefore be overly dependent on the choice of reference model \cite{evans2023bad} and not scale to large compute budgets \cite{goyal2024scaling}.

\textbf{Learnability.} Finally, \citet{mindermann2022prioritized} propose to combine these scores, prioritizing with the difference of losses: $s^\textrm{learn}( \mathcal{B} | \theta, \theta^*) =  s^\textrm{hard}( \mathcal{B} | \theta) + s^\textrm{easy}( \mathcal{B} | \theta^*) =  \ell( \mathcal{B} | \theta) - \ell( \mathcal{B} | \theta^*)$. This heuristic, which we refer to as \textit{learnability} scoring throughout, has the advantage of up-sampling data that is both unlearned and learnable, and has been shown to accelerate large-scale learning even when prioritizing individual examples in isolation \cite{evans2023bad}. In this work, we therefore mainly consider \textit{learnability} scoring but for completeness also provide ablations with \textit{easy reference}  scoring.

The ratio of the ``sub-batch'' and ``super-batch'' sizes defines the \textit{filtering ratio} $f=1 - b/B$, i.e.\ the proportion of data discarded at each iteration. For a given learner batch size $b$, higher filtering ratios increase the cost of scoring as they require more inference passes on the super-batch. % \olivier{Is a sentence like this helpful? Whereas easy-reference scoring can be performed entirely \textit{offline} with a pretrained reference model, both hard learner and learnability scoring depend on the learner's loss and are therefore \textit{online} data curation methods.}

\subsection{Joint example selection (JEST) for multimodal learning}

% \subsection{Multimodal learning losses}
\label{sec:jest}

\textbf{Multimodal learning losses.} Given the availability of internet-scale datasets of paired images and text, multimodal learning has become the default means of training visual representations. Contrastive learning aims to maximize the alignment of these two modalities for paired examples, while minimizing the alignment of unpaired examples.
Both sigmoid- \cite{zhai2023sigmoid} and softmax-contrastive \cite{radford2021learning}
losses achieve this with a batch-level loss $\ell( \mathcal{B} | \theta ) = \frac{1}{b} \sum_{i=1}^b \ell(\vx_i | \theta, \mathcal{B} )$, where the conditional loss $\ell(\vx_i | \theta, \mathcal{B} )$ can use a sigmoid or softmax contrast function (see Appendix Equations \ref{eq:con} and \ref{eq:sig}).
% \begin{align}
% \label{eq:con}
% \ell(\vx_i | \theta, \mathcal{B} ) & = 
% - \frac{1}{2} \left( \log \frac{\exp( \alpha \vz^\textrm{im}_i {\cdot} \vz^\textrm{txt}_i)}{\sum_j \exp( \alpha \vz^\textrm{im}_i {\cdot} \vz^\textrm{txt}_j )}
% + \log \frac{\exp( \alpha \vz^\textrm{im}_i {\cdot} \vz^\textrm{txt}_i)}{\sum_j \exp( \alpha \vz^\textrm{txt}_i {\cdot} \vz^\textrm{im}_j )} \right) \\ 
% \ell(\vx_i | \theta, \mathcal{B} ) & =
% \log \left[ 1 + \exp( - \alpha \vz^\textrm{im}_i {\cdot} \vz^\textrm{txt}_i + \beta ) \right] + \sum_{j \neq i} \log \left[ 1 + \exp( \alpha \vz^\textrm{im}_i {\cdot} \vz^\textrm{txt}_j - \beta ) \right]
% \end{align}
% for the softmax-contrastive \cite{radford2021learning} and sigmoid-contrastive \cite{zhai2023sigmoid} loss, respectively. 
Since \citet{zhai2023sigmoid} demonstrate the sigmoid-contrastive loss to be a more scalable alternative to the softmax-contrastive one, we adopt it by default. Nevertheless, we show in Appendix \ref{sec:softmax} that our results also hold when using the softmax-contrastive loss. % are robust to the exact choice of contrastive loss. %\talfan{we adopt it unless specified otherwise (be more explicit if we include SimCLR results)}. 

\textbf{Joint example selection.}
% \cut{Prior work \cite{evans2023bad,loshchilov2015online,mindermann2022prioritized} has been limited to selecting batches by prioritizing individual examples in isolation.
% While this approach is sensible under the assumption that the loss is independent across examples $\ell( \mathcal{B} | \theta ) = \frac{1}{b} \sum_{i=1}^b \ell(\vx_i | \theta)$
% This approach assumes independence of the loss across examples, which is not the case for the contrastive losses described above As such,} 
% The learnability of a batch jointly depends on the set of examples therein, % \cut{as opposed to simply the sum of each example's learnability. Nevertheless, }
Because the contrastive loss of a batch decomposes into a sum of conditional losses, the \textit{joint learnability} of the batch $s( \mathcal{B} | \theta, \theta^*) \triangleq \ell( \mathcal{B} | \theta) - \ell( \mathcal{B} | \theta^* )  = \frac{1}{b} \sum_{i=1}^b \ell(\vx_i | \theta, \mathcal{B}) - \ell(\vx_i | \theta^*, \mathcal{B}) = \frac{1}{b} \sum_{i=1}^b s(\vx | \theta, \theta^*, \mathcal{B})$
% $s( \mathcal{B} | \theta, \theta^*) = \frac{1}{b} \sum_{i=1}^b s(\vx | \theta, \theta^*, \mathcal{B})$
% \begin{align}
% \label{eq:jointlearn}
% s( \mathcal{B} | \theta, \theta^*) \triangleq \ell( \mathcal{B} | \theta) - \ell( \mathcal{B} | \theta^* )  = \frac{1}{b} \sum_{i=1}^b \ell(\vx_i | \theta, \mathcal{B}) - \ell(\vx_i | \theta^*, \mathcal{B}) = \frac{1}{b} \sum_{i=1}^b s(\vx | \theta, \theta^*, \mathcal{B})
% \end{align}
% In this work we investigate whether jointly prioritizing \textit{learnable batches} is beneficial. %  While prior works have ignored this and independently prioritized examples according to their loss \citet{hessel2021clipscore} or learnability \cite{evans2023bad}, i
%  (see section \ref{sec:obs})
% To do this, we define the \textit{joint learnability} of a batch as 
% \begin{align}
% \label{eq:jointlearn}
% s^\textrm{learn}(\{\vx_i\} | \theta, \theta^*)
% & \triangleq \ell(\{\vx_i\} | \theta) - \ell(\{\vx_i\} | \theta^* ) \\ 
% & = \sum_i \ell(\vx | \theta, \{\vx_j\}) - \ell(\vx | \theta^*, \{\vx_j\}) \\ 
% & = \sum_i s^\textrm{learn}(\vx | \theta, \theta^*, \{\vx_j\}),
% \end{align}
also decomposes into a sum of \textit{conditional learnabilities} $s(\vx | \theta, \theta^*, \mathcal{B})$ of each example given other examples in the batch. 
%We now wish to sample batches proportionally 
We wish to sample batches in proportion to their joint learnability, i.e.\ $p(\{X_k\} = \mathcal{B} ) \propto \exp( s( \mathcal{B} | \theta, \theta^*) )$,
which is enabled by a sequential approach inspired by
%Given its decomposition, % the decomposition above, we follow a sequential approach inspired by 
blocked Gibbs sampling (see Algorithm \ref{alg:sample_sigmoid}). %Specifically, g
Given a subset of examples $\mathcal{B}_n$ already included in the batch at iteration $n$, we compute the conditional learnability of remaining candidate examples $\vx_i$ with $s(\vx_i | \theta, \theta^*, \mathcal{B}_n)$, and sample a new chunk of examples $\{X_k\}$ independently and without replacement according to these probabilities: $p(X_k = \vx_i ) \propto \exp( s( \vx_i | \theta, \theta^*, \mathcal{B}_n) )$. 

We update the batch by appending this chunk to the previous subset: $\mathcal{B}_{n+1} = \mathcal{B}_n \cup \{X_k\}$, and iterate until $n=N$, the number of chunks. The first chunk $\mathcal{B}_{1}$ is sampled using unconditional learnability (i.e. self-similarity only) $s( \vx_i | \theta, \theta^*, \varnothing) ) = \ell(\vx_i | \theta, \varnothing) - \ell(\vx_i | \theta^*, \varnothing)$ where the unconditional losses are computed as $\ell(\vx_i | \theta, \varnothing ) = - \alpha \vz^\textrm{im}_i {\cdot} \vz^\textrm{txt}_i$ for the softmax-contrastive loss and $\ell(\vx_i | \theta, \varnothing ) =
\log \left[ 1 + \exp( - \alpha \vz^\textrm{im}_i {\cdot} \vz^\textrm{txt}_i + \beta ) \right] $ for the sigmoid-contrastive loss.
% \cut{In theory, we could use a chunk-size of 1 (i.e. $N=b$, where $b$ is batch size). However, this would be prohibitively slow in large-scale contrastive learning where  $b=32$k is typical \cite{zhai2023sigmoid}. Nevertheless, }
We find that a relatively small number of chunks ($N=16$, sampling $b/N$ = 2,048 examples independently at each iteration) is sufficient to recover batches with very high learnability (see Section \ref{sec:learn}, Algorithm \ref{alg:sample_sigmoid}).

% \subsection{Online batch selection}
% \label{sec:obs}

% In cases where the loss of a batch is independent across examples, the learnability of a batch is also simply the sum of each example's learnability: 
% % , i.e.\ $\ell(\vx | \theta) = \sum_i \ell(\vx_i | \theta)$
% % $s(\vx | \theta, \theta^*) = \sum_i s(\vx_i | \theta, \theta^*)$
% \begin{equation}
%     \ell(\vx | \theta) = \sum_i \ell(\vx_i | \theta) \implies s(\vx | \theta, \theta^*) = \sum_i s(\vx_i | \theta, \theta^*)
% \end{equation}

% In this case, sampling batches with high learnability simply requires sampling individual examples with high learnability. Specifically, we can use our scoring heuristic to sub-sample a training batch of size $b$ from a larger ``super-batch'' $B$ sampled IID from the training set at each learning update \cite{loshchilov2015online}. Given our scoring model $s(x): \mathbb{R}^{|X|} \rightarrow \mathbb{R}$, each example in the sub-batch is sampled non-uniformly and without replacement in proportion to its score:

% \begin{equation}
% p(X = x_i) \propto \exp( s(x_i) ) 
% \end{equation}
% % , for which one can use online batch selection

\input{Algorithms/joint_example_selection}

% \subsection{Online model approximation}
\subsection{Efficient scoring and multi-resolution training} \label{sec:approx_multires}

% \cut{Large filtering ratios allow JEST to select the most learnable batches, increasing the learning signal per iteration. However, }
\textbf{Efficient scoring with online model approximation.} 
Scoring large super-batches increases the cost per iteration, lowering the efficiency gains in terms of total FLOPs. 
%We consider \textit{model approximation} as a more efficient means of scoring large super-batches. 
While \citet{evans2023bad} required additional small, proxy models to efficiently score data on behalf of larger learners, we remove this requirement by using online model approximation.
%Since the bulk of computational cost of an inference pass comes from the image encoder, we adopt the
We only approximate the image encoding since this accounts for the bulk of the cost of each inference pass \cite{li2023scaling}. 
%instead evaluating the learner at a lower resolution. 
For this we adopt the FlexiViT architecture \cite{beyer2023flexivit}, which lowers the image resolution while minimally degrading performance (see Figure \ref{fig:fig_appendix_multires} for a comparison to patch dropping \cite{li2023scaling}).
% \textbf{Reference score caching.} Since the reference model is pretrained and fixed, its scores can be pre-cached as part of the dataset (see Appendix \ref{sec:ref_cache}), halving the cost of learnability-based scoring. % considerably reducing the cost of scoring.
%Scoring the super batch requires evaluating $B/b$ more examples than when learning from the sub-batch of $b$ images.
%In multimodal learning, the bulk of computational cost lies in encoding images, as the text encoder and contrastive losses are relatively light-weight. 
%We therefore seek to approximate the image encoding of the super-batch. 
%For this we simply 
In our experiments, we evaluate the super-batch with 32$\times$32-pixel patches, 
%as opposed to 16$\times$16, yielding 4$\times$ fewer visual tokens, 
%and dividing the FLOPs 
%per image by 4$\times$. 
%When accounting for the additional overhead of the text encoder and contrastive loss, each example is processed with 28\% of the FLOPs and 33\% of the time of the full-resolution (patch-size 16) model \cite{li2023scaling} (see Section \ref{sec:flop_calcs}).
which gives a 72\% reduction in FLOPs and 67\% reduction in wall-clock time vs. full-resolution scoring at patch size 16$\times$16 \cite{li2023scaling} (see Section \ref{sec:flop_calcs}).

\textbf{Multi-resolution training.} While we want to score examples at low resolution (i.e.\ with large patches), at test-time we wish to evaluate the model at full resolution (i.e.\ with small patches). To enable both resolutions of computation, we simply train at both resolutions. Specifically, given a sub-batch $\mathcal{B}$ for learning, we randomly split it into two halves, $\mathcal{B}^{lo}$ and $\mathcal{B}^{hi}$, and encode each half with a different resolution: $\mathcal{Z}^{lo} = \{ f^\textrm{im}(\vx; \theta, p=32), \vx \in \mathcal{B}^{lo}\}, 
\mathcal{Z}^{hi} = \{ f^\textrm{im}(\vx; \theta, p=16), \vx \in \mathcal{B}^{hi}\}$.
These images embeddings are then concatenated together as $\mathcal{Z} = \mathcal{Z}^{lo} \cup \mathcal{Z}^{hi}$ and the rest of training proceeds as usual. In addition to allowing for efficient scoring, multi-resolution training itself yields a gain in efficiency: since $\mathcal{B}^{lo}$ is processed with 4$\times$ fewer tokens, it also benefits from close to a 4$\times$ speed-up. If $\mathcal{B}^{lo}$ and $\mathcal{B}^{hi}$ each account for half of the batch, the cost of multi-resolution training on $\mathcal{B}$ is 64\% of the FLOPs and 67\% of the time of full-resolution training. Pseudocode for the full-resolution JEST and multi-resolution Flexi-JEST implementation is detailed in Algorithm \ref{alg:training_loop}.

\begin{figure*}
    \centering
    \includegraphics[width=1.0\linewidth]{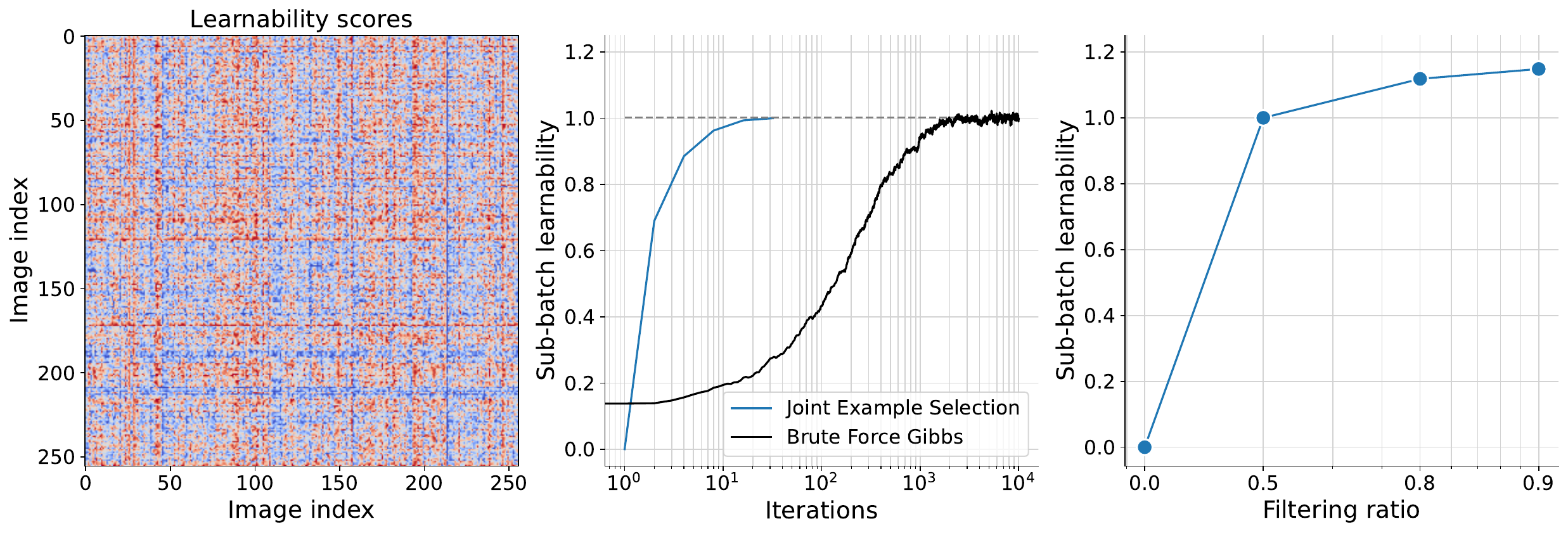}
    \vspace{-1.5em}
    \caption{\textbf{Joint example selection yields more learnable batches. Left:} the learnability of a batch is highly structured and non-diagonal. \textbf{Middle: } Joint example selection quickly discovers sub-batches with high learnability, on-par with brute-force Gibbs sampling. \textbf{Right: } the learnability of sampled batches improves with higher filtering ratios (i.e.\ selecting from larger super-batches). \vspace{-1em}} %  for joint example selection but not independent example selection \cut{i.e.\ scoring and}
    % \cut{strongly depends on pairwise interactions between examples.} 
    \label{fig:fig_pairwise}
    % \talfan{Bigger fonts, resize first panel (try plt.subplots\_adjust(...))} The diagonal is far from capturing this structure. 
\end{figure*}

\textbf{Training datasets.}
% \subsection{Training datasets} 
\label{sec:datasets}
We train the learner model in all JEST experiments on the WebLI dataset \cite{chen2022pali}, specifically 
% We use 
a billion-scale subset of English image-text pairs loosely filtered with % based on a weak 
image-text alignment \cite{zhai2023sigmoid}, using the \texttt{big\_vision} codebase \citep{big_vision}. To train reference models, we use smaller high-quality datasets. JEST/Flexi-JEST reference models are trained on a strongly filtered 100M scale subset of WebLI filtered for high text and image quality and image-text alignment, which we refer to as ``WebLI-curated''. We additionally explore 
\textit{scaling data curation} (JEST++/FlexiJEST++) using reference models trained on ``WebLI-curated++'' which adds approximately 600M additional web-scraped image-text pairs
%(approx. 600M examples) 
filtered with the same strong curation pipeline. Unless specified otherwise, the average performance we report uses 4 canonical benchmarks for multimodal transfer: %all average performance metrics are reported on 4 commonly-used benchmarks: 
ImageNet 0-Shot and 10-Shot classification and COCO image-to-text and text-to-image top-1 retrieval.
% For more details on the impact of different curated datasets see Sec. \ref{sec:curation}. 

%% file: Algorithms/joint_example_selection.tex
\begin{algorithm}[t]
\caption{Joint example selection: sigmoid loss}
\label{alg:sample_sigmoid}
\definecolor{codeblue}{rgb}{0.25,0.5,0.5}
\definecolor{codegreen}{rgb}{0.25,0.5,0.25}
\lstset{
  backgroundcolor=\color{ouralg!20},
  basicstyle=\fontsize{7.2pt}{7.2pt}\ttfamily\selectfont,
  columns=fullflexible,
  breaklines=true,
  captionpos=b,
  commentstyle=\fontsize{7.2pt}{7.2pt}\color{codeblue},
  keywordstyle=\fontsize{7.2pt}{7.2pt}\color{codeblue},  % \color{codeblue}
}

\lstset{keepspaces=true}
\begin{lstlisting}[language=python]
def jointly_sample_batch(learner_loss, ref_loss, n_chunks=16, filter_ratio=0.8, method="learnability"):
  scores = learner_loss - ref_loss if method == "learnability" else - ref_loss
  n_images = scores.shape[0]                               # scores.shape = [B, B]
  n_draws = int(n_images * (1 - filter_ratio) / n_chunks)  # Size of each chunk.
  logits_ii = np.diag(scores)                              # Self-similarity scores.
  inds = random.choice(logits_ii, n_draws)                 # Sample first chunk.

  for _ in range(n_chunks - 1):
    is_sampled = np.eye(n_images)[inds].sum(axis=0) # Binary indicator of current samples [n_images,].
    logits_ij = (scores * is_sampled.reshape(n_images, 1)).sum(axis=0) # Negative terms ij [n_images,].
    logits_ji = (scores * is_sampled.reshape(1, n_images)).sum(axis=1) # Negative terms ji [n_images,].
    logits = logits_ii + logits_ij + logits_ji      # Conditional learnability given past samples.
    logits = logits - is_sampled * 1e8              # Avoid sampling with replacement.
    new_inds = random.choice(n_images, n_draws, p=np.exp(logits))
    inds = np.concatenate((inds, new_inds))         # Expand the array of indices sampled.
  return inds                                       # Gather and return subset indices.
\end{lstlisting}
% \vspace{-1em}
\end{algorithm}

%% file: Sections/experiments.tex
\section{Experiments}

\subsection{Joint example selection yields learnable batches} 
\label{sec:learn}

We start  by evaluating the efficacy of joint example selection (JEST) for selecting learnable batches. To gain an intuition for our method, we start by visualizing the learnability matrix (i.e.\ the difference in loss between learner and reference models, for all pairs of examples in the batch).
%for a small batch of examples
%early on in training. 
%As expected, the total learnability (i.e.\ the sum of the learnability matrix) depends on much more than its diagonal terms only, as the matrix is highly structured and non-diagonal (Figure \ref{fig:fig_pairwise}, left).
JEST is designed to sample sub-matrices of examples in proportion to their summed learnability. Since the matrix is strongly non-diagonal (Figure \ref{fig:fig_pairwise}, left), independent selection will clearly be sub-optimal. % (as in e.g.\ CLIP-Score)

% \cut{We next assess the ability of JEST to extract highly learnable sub-batches from the super-batch.} 
With a small number of iterations (corresponding to populating the batch with $N=16$ chunks), we find the learnability of the sub-batch to quickly increase, matching the learnability of batches extracted by brute-force Gibbs sampling requiring thousands of iterations (Figure \ref{fig:fig_pairwise}, middle).

% \cut{Finally, we assess the increase in learnability afforded by selecting from increasingly large super-batches.} 
%To arrive at f
For filtering ratios of 0.5, 0.8, and 0.9, we select sub-batches of 32,768 examples from super-batches of size 65,536, 163,840 and 327,680 respectively. In Figure \ref{fig:fig_pairwise}, right, we find that the learnability of the sub-batch increases with larger filtering ratios.
%, as it has a larger super-batch to choose from. %, a property missing from the more standard ``positive-only'' selection method used in prior work (which we arrive at by setting $N=1$ chunk in our algorithm). 
In summary, our joint example selection (JEST) algorithm is an effective and efficient means of selecting highly learnable batches during training. % , whose affect on performance 

% \begin{enumerate}
%     \item Previous data curation methods evaluate individual data ahead of training.
%     \item However, for contrastive training, the relevance of any data point for training cannot be computed independently of the other data in a given batch, since the off-diagonal (negative) terms of the contrastive matrix contribute significantly to the overall loss (Figure \ref{fig:fig_pairwise}).
%     \item In fact, negatives may contribute most of the mass of the learnability matrix (i.e.\ the difference in pairwise loss between the learner and reference model. Indeed, visualizing this matrix for 256 randomly sampled data points shows that it is strongly non-diagonally recessive (see Figure \ref{fig:fig_pairwise}, left).
%     \item Instead, data must be selected jointly such as to produce the maximal interaction. If we are rejecting 50\% of the data for a given batch, this is equivalent to selecting from the $2N \times 2N$ matrix the $N \times N$ sub-matrix which has the largest summed entries.
%     \item To do this, we devised a sequential strategy based on blocked Gibbs-Sampling, which we call Joint Example Selection (JEST).
%     \item Blocked Gibbs-Sampling (\textcolor{red}{olivier: See Wiki}) is much more efficient and performs just as well as full Gibbs sampling in our experiments. This is a negligible overhead vs. the costs of training.
%     \item We show that JEST outperforms independent sampling (positive-only, if selecting only based on the diagonals of the contrastive matrix) across all metrics.
% \end{enumerate}

\begin{figure*}
    \centering
    \includegraphics[width=1.0\linewidth]{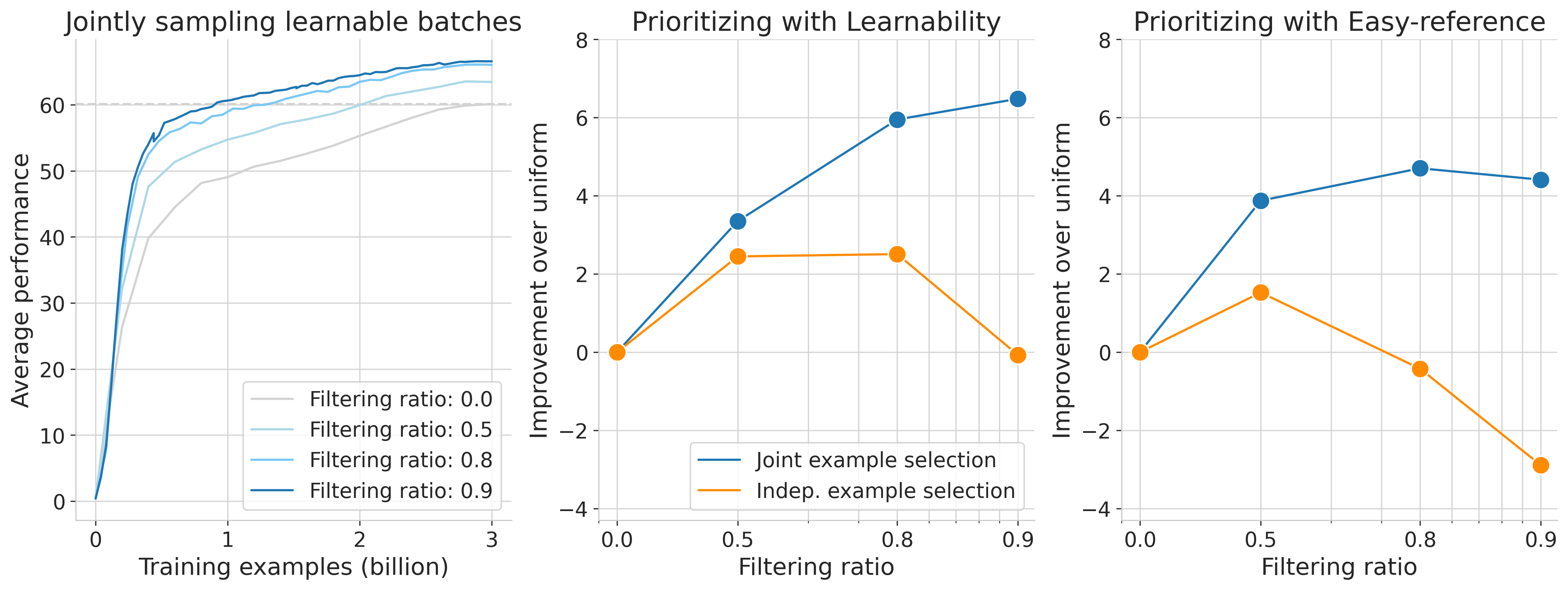}
    \vspace{-1.5em}
    \caption{\textbf{Joint example selection accelerates multimodal learning. Left:} training on the most learnable sub-batch selected from super-batches that are 2$\times$, 5$\times$, or 10$\times$ larger significantly accelerates multimodal learning. \textbf{Middle: } Jointly prioritizing \textit{learnable batches} yields significantly better results than simply prioritizing individual examples. \textbf{Right: } joint examples selection also improves \textit{easy reference} prioritization, although \textit{learnability} scales better with more aggressive filtering.
    \vspace{-1em}
    }
    \label{fig:fig_spi}
\end{figure*}

\subsection{Joint example selection accelerates multimodal learning}
\label{sec:spi}

% learnability-based prioritization with 
% \olivier{Clarify training details.}
We now investigate the effect of training on more learnable batches, as selected by our JEST algorithm. All runs use a reference model trained on WebLI-curated, % small, curated dataset (see section \ref{sec:curation})
 a ViT-B/16 and Bert-B image-text dual encoder, 3 billion training examples, and the sigmoid-contrastive loss. Figure \ref{fig:fig_spi} (left) shows the average performance on multiple downstream tasks (ImageNet 0-Shot/10-Shot accuracy and COCO image-to-text/text-to-image retrieval) over the course of training. %, for models trained with 0\% filtering (i.e. standard uniform training) up to 90\% filtering. 
 We find that JEST 
 significantly accelerates learning,
%  yields significant improvements to both final performance and the rate of convergence, particularly with high filtering ratios. In terms of acceleration, JEST-trained models 
 reaching the final performance of the 3B-uniform baseline after only 2B, 1B, and 0.67B training examples, when using filtering ratios of 50\%, 80\%, and 90\% respectively. % with increasingly large filtering ratios. 
 At larger filtering ratios we observe similar training instabilities to those observed for larger batch sizes \cite{zhai2023sigmoid}, necessitating a modification to stabilize the Adam optimizer ($\beta_2 = 0.95$) and suggesting that data curation with JEST can be thought of as increasing the effective batch size (Appendix \ref{sec:beta_opt}, \ref{sec:effective_batch_size}).  % Appendix
% , for each model
% \talfan{Add refrence to effective batch size experiments.}
% that does not affect the IID runs

In terms of final performance, JEST also delivers significant gains of up to
%3.5\% given 50\% filtering and over up to 
6\% 
%with more aggressive filtering 
when filtering 90\% of data (Figure \ref{fig:fig_spi}, middle, blue curve). Notably, this scaling behavior is absent from previous selection methods based on independent prioritization of individual examples (Figure \ref{fig:fig_spi}, middle, orange curve). %Whereas moderate filtering (around 50\% or 80\%) yields some improvement (up to 2\%, consistently with \citet{evans2023bad}), these improvements do not scale: filtering more aggressively yields a regression in performance (Figure \ref{fig:fig_spi}, middle, orange curve)
Finally, we assess whether 
%the JEST framework 
JEST also improves prioritization criteria other than learnability. Figure \ref{fig:fig_spi}, right, shows the performance of models with \textit{easy-reference} prioritization, for varying filtering ratios. Consistent with learnability-based prioritization, JEST strongly outperforms independent example selection, particularly for high filtering ratios (where independent example selection leads to a regression in performance). Prioritising data with the highest loss produced smaller gains and degrades more quickly as we filter more data (Figure \ref{fig:tigg_hard_learner}). Since learnability-based JEST yielded the best scaling behavior 
%than \textit{easy-reference} JEST 
we retain this criterion for subsequent experiments. 
% \cut{In summary, by selecting more learnable batches, JEST significantly accelerates and improves multimodal pretraining.}
% relative to the uniform baseline

\subsection{Synergies between multi-resolution training and online batch selection}
\begin{figure*}[h]
    \centering
    \includegraphics[width=1.0\linewidth]{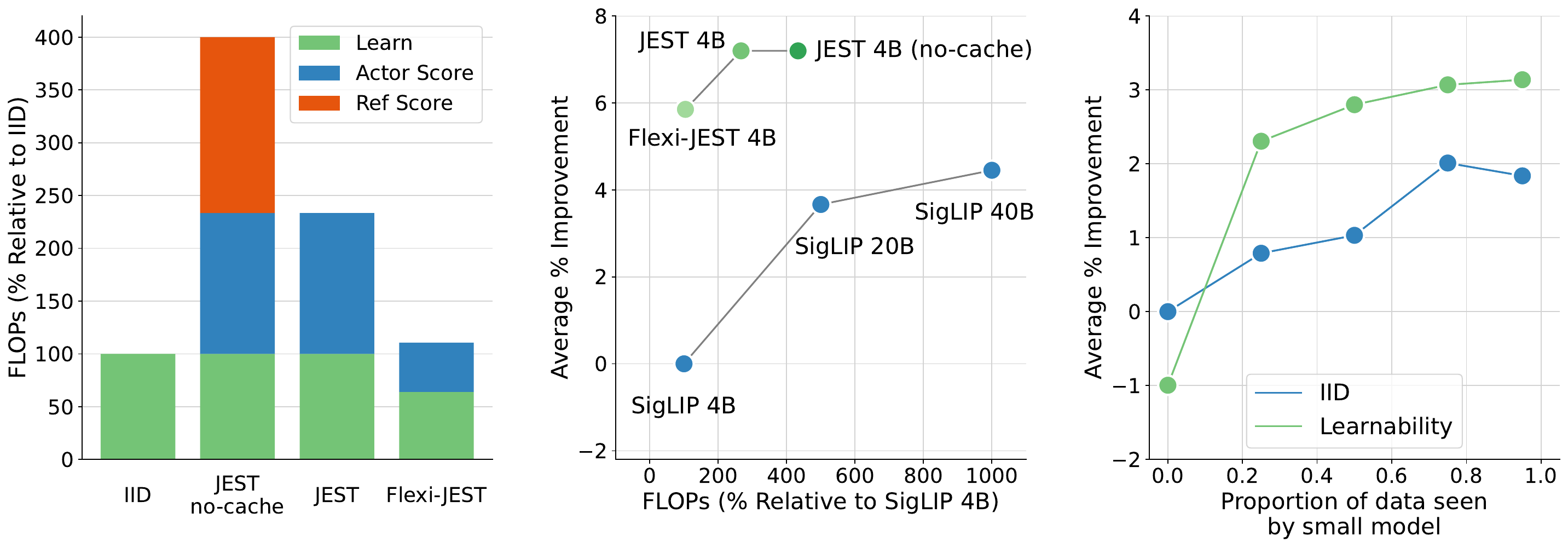}
    \vspace{-1.5em}
    \caption{\textbf{Efficient scoring and multi-resolution training }.
    % Approximating the actor model (by increasing patch size) is desirable to reduce FLOPs, but scoring without training the approximate model can result in poor performance. Inspired by \cite{beyer2023flexivit, li2023scaling}, we demonstrate that \textit{co-training} both the full and approximate learner ameliorates this out-of-domain gap. 
    \textbf{Left:} 
    In scoring large super-batches with the learner and reference models, JEST incurs a large computational cost per iteration. %(roughly 4$\times$ that of standard IID training).
    % Illustration of the costs of each operation in each JEST variant described in Section \ref{sec:approx_multires}. 
    By caching the fixed reference model scores in the dataset, this overhead can be cut in half. 
    % ``Cache Ref.'' refers to the cost saving associated with storing the fixed reference model scores in the dataset. 
    Efficient scoring and multi-resolution training further reduce this to be comparable to standard IID training. 
    \textbf{Middle: } Flexi-JEST improves the total FLOP-efficiency of JEST over standard IID training. % JEST and Flexi-JEST trade off FLOP efficiency with data efficiency, but both are superior to IID training (also see Figure \ref{fig:fig_speedups_teaser}). 
    \textbf{Right: } Multi-resolution training improves FlexiJEST more than standard IID training. 
    % Improving approximate scoring performance for Flexi-JEST by co-training the full and approximate learner models. In this experiment, 
    % we vary the proportion of data used for low-resolution training (i.e.\ with patch size 32) and match for overall training compute by adjusting the training schedule accordingly. 
    Without multi-resolution training (left-most point) Flexi-JEST underperforms the IID baseline (due to an untrained approximate model), but quickly improves with even a small amount of co-training (25\%). \vspace{-1em}
    } 
    \label{fig:fig_multires}
\end{figure*}

Joint example selection 
with \textit{learnability} scores becomes more efficient as larger fractions of each batch are filtered.
%greatly improves multimodal pretraining. %
% allows us to extract greater
%enables great
%training efficiencies by filtering large fractions of each batch. 
However, the cost of scoring 
%the super-batch 
results in a significant overhead:
%a naive implementation of JEST 
filtering 80\% of the super-batch results in 4$\times$ more FLOPs per iteration than IID training, or 2.3$\times$ when caching the reference-model scores (Appendix \ref{sec:flop_calcs}). 
Although JEST is significantly more efficient in terms of training iterations (hereinafter `training efficiency'), the additional scoring FLOPs reduce its compute efficiency relative to the IID baseline
%compared to training-efficiency 
(Figure \ref{fig:fig_speedups_teaser}, left vs. right). We therefore %introduce 
also investigated a compute efficient variant, Flexi-JEST, which uses multi-resolution training and low-resolution scoring to 
% (which are fixed over the course of training) allow us to reduce this cost to 2.33$\times$ that of IID training. 
% Finally, making both training and scoring more efficient through multi-resolution training and low-resolution scoring 
reduce the total overhead to only 10\% vs. the baseline 
(Figure \ref{fig:fig_multires}, left; see Section \ref{sec:flop_calcs}).     %   increases with the filtering ratio, as the size of the super-batch \cut{However, the total scoring cost scales inversely with the filtering ratio, since each rejected data point requires an inference pass from the learner. This means that each JEST iteration requires 2.33x as many FLOPs as a comparable IID iteration at a filtering ratio of 0.2 (rejecting 80\% of data; see Appendix Section \ref{sec:flop_calcs})}. 
%  for details

What is the effect of these approximations on performance? As might be expected, the per-iteration performance of Flexi-JEST decreases relative to the JEST, although still produces significant speed-ups over IID (Figure \ref{fig:fig_speedups_teaser}, left; Figure \ref{fig:fig_multires}, middle). However, the decrease in per-iteration performance is more than favorable when accounting for the decrease in total FLOPS: our best Flexi-JEST model produces the same average performance as a 40B Siglip run with 9.9$\times$ fewer FLOPs, and ~2$\times$ fewer than full-resolution JEST  (Figure \ref{fig:fig_speedups_teaser}, right; Figure \ref{fig:fig_multires}, middle). % \talfan{Maybe save this for comparison to prior work section.}

What is the relative contribution of efficient scoring and multi-resolution training in Flexi-JEST? 
% \cut{Previous work (\cite{evans2023bad} has demonstrated that much smaller models can still be used as effective proxies for scoring data for training of much larger models.} 
% We therefore investigated the use of model approximation via patch resizing \cite{beyer2023flexivit} described in Section \ref{sec:approx_multires} \cut{to reduce the cost of the data scoring. We use a patch size of 32, which is a reduction of 4x relative to the learner. In real terms, this corresponds to an approximately 72\% reduction in FLOPs per forward pass \cite{li2023scaling} since we do not drop tokens on the learner.}
% \cut{Our initial experiment showed a surprising reduction in performance vs. IID (Figure \ref{fig:fig_multires}, Right). We hypothesized that this reduction could be explained by the scoring model operating out-of-distribution, since only the full learner at patch size 32 was actually being trained. }To ameliorate this 
We conducted an ablation where we varied the fraction of the selected batch trained at full and low resolution (i.e.\ the relative sizes of $\mathcal{B}^{hi}$ and $\mathcal{B}^{lo}$; see Methods). 
%For fair comparison we 
We ensure the learner spends the same FLOPs by increasing the number of training iterations as we send more data to the approximate model, since the FLOPs per iteration decrease in this case (see Section \ref{sec:app-approximation} for details). %\olivier{Detail this calculation in the appendix?}  % by the full and approximate learner
Figure \ref{fig:fig_multires} (right) shows that the IID baseline performance increases with larger fractions of data sent to the approximate model, consistent with a growing literature on the FLOP-efficiency of approximate training \cite{beyer2023flexivit, li2023scaling,dehghani2024patch,raposo2024mixture}.
% Confirming our hypothesis, 
Nevertheless, Flexi-JEST significantly outperforms the multi-resolution baseline as long as the low-res model trains on enough data to %align with the learner and so 
align with the learner
%as it trains the approximate model
(e.g. $\geq$ 25\% data). %  used for approximate training 
% performance quickly exceeds the IID baseline, as we co-train the full and approximate learner models, significantly reducing the OOD gap with a small amount (25\%) of training. 
% Although isoFLOP performance continues to increase as we send more data to the small model, we choose a 50\% split as our canonical Flexi-JEST implementation going forward.
% Indeed given that scoring is performed at low-resolution, JEST 
These experiments demonstrate a synergy between multi-resolution training and joint example selection, as the former yields efficient and accurate scoring capabilities for accelerating the latter. % scoring  active learning. % A chart of the FLOP contributions is shown in Figure \ref{fig:fig_multires}.
% \talfan{Why remove ref to bar chart?} \olivier{Already discussing this in 1st paragraph}

Our results also point to a pareto front of data curation strategies. If maximizing training speed or training efficiency is desirable at the expense of computation, the full-resolution JEST method produces up to a 13$\times$ speed up relative to a comparable IID training run. If FLOPs should be minimized at the expense of training efficiency, Flexi-JEST produces the most favorable trade-off. % performance
We note that the scoring of the next batch can be implemented on separate devices, in parallel with training, 
%representing an alternative to scaling computation via batch size increases.
potentially further reducing the additional wall-clock time.
%to zero.

% \subsection{Two-stage filtering further reduces scoring costs \textcolor{red}{TODO(talfan)}}

% \begin{enumerate}
%     \item Scoring with approximate models is much cheaper, but is slightly less performant than scoring with non-approximate models.
%     \item Can we get the best of both worlds?
%     \item We propose to perform filtering in two stages: once with a signifiacntly cheaper model (100\% $\rightarrow$ 40\%), the secondly with the full model (40\% $\rightarrow$ 20\%).
%     \item \textcolor{red}{We have initial results but full runs TBD, this is non-essential to the story at this point.}
% \end{enumerate}

% \subsection{Making sense of data curation for active learning}
\subsection{Joint examples selection enables strong data-quality bootstrapping}

\label{sec:curation}
% \cut{An essential component of any learnability-based active learning method (including JEST) is the choice of reference model. Assuming the reference model is of similar architecture to the learner, the critical variable then becomes the reference training dataset. \cite{evans2023bad} show preliminary evidence that training reference models on curated subsets (of the WebLi dataset) does in fact improve active learning performance, yet many questions around this topic were left unanswered.}

At the heart of learnability-based scoring is a reference model trained on a small, curated dataset of our choosing. 
% In this section we ask two questions: 
How does JEST performance vary as a function of different curation strategies that trade off quality vs. quantity? Furthermore, do improvements in JEST training correlate with the performance of the reference models or are these metrics decoupled?   %  performance

% that vary on a spectrum from weak to strong
\textbf{Understanding quality vs. quantity trade-offs}. We explore three scales of curation, each being a subset of the original WebLI dataset: 
\textit{weak} (billion-scale) curation 
% \textit{Weak (billion-scale)}: weakly curated 
with image-text alignment (ITA) filters, 
\textit{moderate} (300M scale) curation with either ITA filters or text-quality (TQ) filters, and \textit{strong} (100M scale) curation with a combination of TQ, ITA, and additional image-quality (aesthetic) filters. 
% . \textit{Moderate (300M scale)}: two moderately curated subsets either filtered with the same ITA filters or text-quality (TQ) filters. 
% \textit{Strong (100M scale)}: strongly curated with a combination of TQ, ITA, and additional image-quality (aesthetic) filters.
Throughout, we refer to this strongly curated subset as ``WebLI-curated''. 

\begin{figure*}
    \centering
    \includegraphics[width=0.9\linewidth]{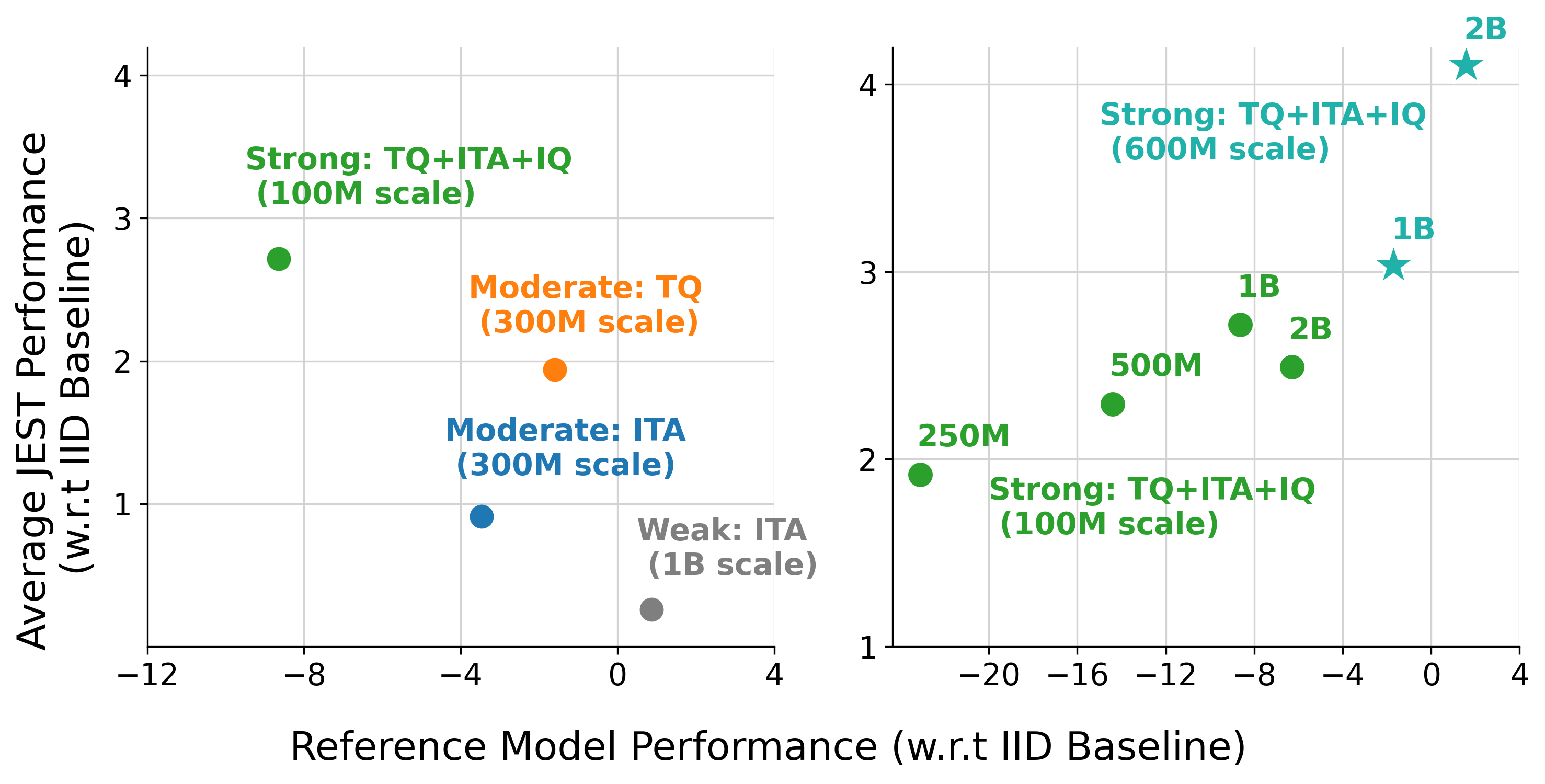}
    \caption{\textbf{Scaling strong data curation improves JEST performance.} \textbf{Left:}  We compare JEST performance vs.\ reference model performance (relative to the uniform baseline) for 4 curation types: `weak' curation with image-text alignment (ITA), `moderate' curation with ITA or text-quality (TQ), and  % and image-text alignment (ITA),  %show strong correlation between reference performance and JEST performance. 
    %  (ITA filters) or 
     `strong' curation (using a combination of TQ, ITA, and additional image-quality (IQ). % filters, green points), leads to the best JEST model and a decoupling between reference and JEST performance. 
     \textbf{Right:} We use our best reference dataset (TQ+ITA+IQ) and evaluate JEST vs. reference performance varying the number of examples seen during reference pretraining. There is a strong correlation between additional reference training and JEST performance that saturates after 1B examples seen. By scaling strong data curation to a 600M dataset, this saturation is broken as both reference model and JEST performance improve for the 1B \textit{and} 2B reference training. 
    %  \olivier{Looks great! Could you move the "Strong: TQ+ITA+IQ" label a bit to the right?} 
     \vspace{-1em} 
    } 
    \label{fig:fig_curation}
\end{figure*}

We train standard SigLIP encoders on these four WebLI subsets for 10 epochs each, and use them as reference models for JEST training on the full WebLI dataset. % followed by JEST runs using these reference models for 3 pretraining data epochs. 
Across curation methods, reference model performance and JEST performance appear to be decoupled (or even anti-correlated; Figure \ref{fig:fig_curation}, left), consistent with % ($\Delta$ w.r.t IID baseline). 
% This is similar to 
previous findings for fixed data curation \cite{fang2023data}. 
Whereas increasing curation (and decreasing dataset size) yields weaker models, when used as reference models for JEST pretraining they have the opposite effect: JEST with a strongly-curated reference benefits from a 2.7\% improvement, moderate a 1.5\% improvement, and weak a 0.3\% improvement. % (averaged over TQ and ITA)

\textbf{Scaling data curation}.
%While the results in Fig. \ref{fig:fig_curation} (Left) suggest that there is an inherent decoupling between reference performance and JEST performance for highly-curated data, w
We hypothesized that 
the general decoupling between reference model performance and JEST performance might simply be explained by the dataset size limits imposed by data curation.
%this may simply be due to the fact that stronger curation limits total dataset size. 
To understand this effect, we trained 5 reference models on WebLI-curated while varying the total examples seen (250M to 3B). In this context, Figure \ref{fig:fig_curation} (right) shows a striking correlation between improved reference models and better JEST pretraining. This suggests that the ``decoupling'' phenomenon can be mostly attributed to the saturation of the reference model as a result of the reductions in dataset size following curation.

We note that the correlation in Figure \ref{fig:fig_curation} (right) starts to break down when the dataset is saturated, i.e.\ after 10 epochs or  1B examples seen. 
These results suggest that JEST would benefit further from scaling data curation of the reference datasets. To test this, we grew WebLI-curated to approximately 600M examples sourced from an expanded set of image-text pairs. At this scale, however, it is difficult to satisfy the stringent TQ/ITA criteria of WebLI-curated. Therefore, for all image-text pairs that did not meet the original ITA threshold, we re-captioned the images with high-quality synthetic captions based on the PaLI model family \cite{chen2022pali, chen2023pali}---we denote this dataset as ``WebLI-curated++''. We find that this scaled dataset allows us to break the 2B saturation point for ``WebLI-curated'' as both reference model and JEST performance (Figure \ref{fig:fig_curation}, Right: $\color{cyan} \bm{\star}$) improves significantly. We therefore use WebLI-curated++ for our best models, JEST++ and FlexiJEST++. % In fact, to achieve our best JEST++ result, we find further improvements by training the reference model for a total of 5B examples.
 %by growing WebLI-curated to a total of approximately 600M examples sourced from an expanded set of image-text pairs. In order to satisfy the strict criteria on ITA,

Given that scaling data curation with WebLI-curated++ strongly improves reference model performance, we asked whether pretraining on the original WebLI dataset is necessary at all. However when evaluating the reference model's performance \textit{across datasets} we find it to be very imbalanced: while it outperforms WebLI pretraining on 2 downstream tasks, it significantly underperforms on 6 others, as well as on average (Table \ref{tab:appendix_table_2}). In contrast, JEST++ pretraining on WebLI yields a \textit{generalist} foundation model that outperforms the baseline on 6 of 8 benchmarks, as well as on average. 

% On closer inspection, we show that the reference model is in fact very specialized in performance (Table \ref{tab:appendix_table_2}). Indeed, JEST++ pretraining is necessary to leverage this specialized reference model for training a \textit{generalist} foundation model.

\input{Tables/main_table}

\subsection{Comparison to prior art}

We now compare to prior art, including the state-of-art SigLIP model trained for 40 billion examples \cite{zhai2023sigmoid} as well as recent strong CLIP variants. 
% We evaluate zero- and few-shot performance across 8 downstream tasks. 
Table \ref{tab:main_table} shows that our most training-efficient model, JEST++, sets a new state-of-the-art on both ImageNet and COCO all while using 10$\times$ fewer iterations and 4$\times$ less compute. On COCO in particular, JEST++ improves the previous state of the art by over 5\%. Our most compute-efficient model, Flexi-JEST++, also surpasses the previous SoTA on average, while using 9$\times$ less compute. Training JEST for longer furthered these gains (see Appendix Table \ref{tab:appendix_table}). %Furthermore, our most compute-efficient model, FlexiJEST++, still outperforms SigLIP (40B) on average while requiring only 11 \% of the total FLOPS.
%  when averaging across metrics and in 6 out of 8 evaluations taken individually,
% Longer training runs produced further improvements

Our results also scale gracefully with model size. Training with a ViT-L learner and ViT-L reference trained on the same WebLI-curated++ dataset, 
%We next trained JEST++ with a ViT-L architecture, using a ViT-L reference model trained on Webli-curated++. 
%shows that 
JEST++ 
%also yields 
continues to yield strongly accelerated learning,
%at this scale, 
matching the SigLIP ViT-L 40B baseline with only 4B examples seen (Table \ref{tab:main_table}, bottom). 

% asked whether the accelerated learning afforded by JEST holds across scale, and trained a ViT-L reference model on Webli-curated++, and used it to train a ViT-L on Webli with JEST++. 

Finally, we apply JEST++ for pretraining on the publicly available LAION-2B dataset \cite{schuhmann2022laion}. We follow the standard practice of removing unsafe image-text pairs \cite{singer2022make}, but do not otherwise pre-filter the dataset. JEST++ strongly surpasses previous methods for offline data curation, despite using 4$\times$ fewer training examples than the previous state-of-the-art (Table \ref{tab:test}). With this training budget, SigLIP pretraining severely under-performs all methods, further highlighting the benefits of JEST. %  (e.g.  fewer) showing improved performance over all methods while requiring significantly fewer training examples.
% (WebLi-curated++ reference model) 
%  curate
% the efficiency gains achieved by
% (e.g.\ +1.5\% and +6.3\% on ImageNet and COCO)

\input{Tables/laion_table}

\subsection{Simplifying data curation}
%\textbf{Removing the need for offline curation}.
%
% Here we demonstrate that with a strong reference model, the JEST++ methodology simplifies the data preparation process allowing for pretraining on completely uncurated datasets. The original WebLI dataset \cite{chen2022pali} that we use for all pretraining experiments is in fact filtered for high image-text alignment and this off-line pre-filtering is common for preparing web-scale image-text datasets. In Fig. \ref{fig:fig_rawvsfiltered}, we demonstrate that JEST++ eliminates the need for any filtering of pretraining datasets, greatly simplifying dataset preparation.
%
%With a strong reference model, the JEST++ methodology simplifies the data preparation process allowing for pretraining on completely uncurated datasets. 
The WebLI dataset \cite{chen2022pali} used for our pretraining experiments has already been filtered for high image-text alignment. As shown in Table \ref{tab:rawvsfilt}, this offline curation is critical for strong IID training performance. For JEST++ however, this pre-filtering is redundant as performance does not degrade when training on the unfiltered WebLI, alleviating the need for foundation datasets. 

\input{Tables/rawvsfiltered}

% \begin{figure*}[t]
%     \centering
%     \includegraphics[width=0.8\linewidth]{Figures/fig_7_jest_rawvsi18n.png}
%     \caption{\textbf{Simplifying data curation.} JEST++ simplifies the data curation process by allowing for pretraining on completely uncurated datasets. \textbf{Left:} With uniform (IID) pretraining, there is normally a large gap between pretraining on raw image-text data and data pre-filtered for high image-text alignment. JEST++ largely closes this gap. \textbf{Right:} In fact, the final performance for JEST++ is nearly identical for pretraining on raw vs. filtered data (standard deviation across benchmarks is shown). Performance is shown relative to the IID (filtered) baseline. }
%     \label{fig:fig_rawvsfiltered}
% \end{figure*}

%% file: Tables/main_table.tex
\def\tbd{\textcolor{red}{TBD}}

% flops_per_image_siglip_B_16 = 35.1 * 3
% flops_per_image_jest_B_16 = 35.1 * (2 + 1 / SPI) if we cache the inference pass
%
% ratio = 7/3 = 2.333 for SPI = 0.2
%
% (3 * (0.5 * 1.0 + 0.5 * 0.25)) + 1 / SPI * 0.25
% = 3 * 0.625 + 0.25 * 5
% = 3.125
%
% = 1.042, so 4.2% overhead

\begin{table*}[h]
\centering
%\footnotesize
% \scriptsize
%\footscriptsize
\small

\newcommand*\rot{\rotatebox{90}}

% Paste output from Colab:
%   https://colab.corp.google.com/drive/1jYI-LEGIaf0gL5RTHff0IQwYVeu_m9MW?usp=sharing

% \definecolor{goodgreen}{rgb}{0.25,0.7,0.25}
% \definecolor{goodred}{rgb}{0.9,0.25,0.25}
\definecolor{goodgreen}{rgb}{1.0,1.0,1.0}
\definecolor{goodred}{rgb}{1.0,1.0,1.0}

\newcommand{\coloredcell}[3]{%
    \pgfmathsetmacro{\maxval}{#2}%
    \pgfmathsetmacro{\minval}{#3}%
    \pgfmathsetmacro{\scalar}{#1}%
    \pgfmathsetmacro{\range}{\maxval - \minval}%
    \pgfmathsetmacro{\normscalar}{(\scalar - \minval) / (\maxval - \minval) * 2 - 1}%
    \pgfmathsetmacro{\opacity}{100 * abs(\normscalar) * 0.8}%
    \ifdim\normscalar pt > 0pt
        \edef\tempcell{\noexpand\cellcolor{goodgreen!\opacity!white}}%
        \tempcell \scalar%
    \else
        \edef\tempcell{\noexpand\cellcolor{goodred!\opacity!white}}%
        \tempcell \scalar%
    \fi
}

\newcommand{\coloredcellpositive}[3]{%
    \pgfmathsetmacro{\textscalar}{#1}%
    \pgfmathsetmacro{\maxval}{#2}%
    \pgfmathsetmacro{\valscalar}{#3}%
    \pgfmathsetmacro{\normscalar}{\valscalar}%
    \pgfmathsetmacro{\opacity}{100 * log10(\valscalar) * 0.8}%
    \edef\tempcell{\noexpand\cellcolor{goodgreen!\opacity!white}}%
    \tempcell \textscalar%
}

\newcommand{\coloredcellnegative}[3]{%
    \pgfmathsetmacro{\textscalar}{#1}%
    \pgfmathsetmacro{\maxval}{#2}%
    \pgfmathsetmacro{\valscalar}{#3}%
    \pgfmathsetmacro{\normscalar}{\valscalar}%
    \pgfmathsetmacro{\opacity}{100 * log10(\valscalar) * 0.8}%
    \edef\tempcell{\noexpand\cellcolor{goodred!\opacity!white}}%
    \tempcell \textscalar%
}

% Paste here

% \resizebox{\linewidth}{!}{

\begin{tabular}{lcccc|ccccc}
%\begin{tabular}{@{\extracolsep{1pt}}lccccccccccc|c@{}}

& & & \multicolumn{2}{c}{FLOPs \%} & 
%\textbf{Mean} 
& \multicolumn{2}{c}{ImageNet-1K} & \multicolumn{2}{c}{COCO} \\ % First row
\cmidrule(lr){4-5} \cmidrule(lr){7-8} \cmidrule(lr){9-10}
% \cline{4-5} \cline{7-8} \cline{9-10}
\\ [-2.4ex]
%\noalign{\smallskip}
Method & Variant & \# Train & Per Iter. & \textbf{Total} & \textbf{Mean} $\Delta$ & 10-S & ZS &  I2T & T2I % & Birds & Caltech &  Cars & Pets \\% & $\mathbf{\Delta}$ \\  % Second row
\\

% Method&\rot{FLOPs (\% Per Iter.)}&\rot{FLOPs (\% Total)}&\rot{Train Ex. (B)}&\rot{IN 10-Shot}&\rot{IN-1K}&\rot{COCO I2T R1}&\rot{COCO T2I R1}&\rot{Birds 10-Shot}&\rot{Caltech 10-Shot}&\rot{Cars 10-Shot}&\rot{Pets 10-Shot}&\rot{Mean}&$\Delta$ (\%)\\

\hline
% \cmidrule(lr){1-10}

CLIP \cite{radford2021learning}&B&13B&       100&\hspace{0.5em}32&-- 11.8&&68.3&52.4&33.1\\
EVA-CLIP \cite{sun2023eva}&B&\hspace{0.5em}8B&       100&\hspace{0.5em}20&\hspace{0.4em}-- 4.6&&74.7&58.7&42.2\\
OpenCLIP \cite{ilharco_gabriel_2021_5143773}&B&34B&       100&\hspace{0.5em}85&\hspace{0.4em}-- 5.8&&70.2&59.4&42.3\\
LessIsMore \cite{cao2023less}&B&11B&       100&\hspace{0.5em}28&\hspace{0.4em}-- 5.9&&70.8&58.3&42.5\\
SILC-S \cite{naeem2023silc}&B&20B&       380&190&\hspace{0.1em}$+$ 0.2&68.9&76.6&66.2&48.7\\
% &&&&&&&&&\\[-0.5em]
SigLIP \cite{zhai2023sigmoid}&B&40B&       100&100&\hspace{1.1em}0.0&70.3&76.7&65.2&47.4\\

% \rowcolor{DnCBG} JEST&B&\hspace{0.5em}5B&       233&29&-0.5&68.2&75.5&66.1&47.9\\
% \rowcolor{DnCBG} Flexi-JEST&B&13B&       110&36&1.2&69.4&76.6&68.2&50.2\\
\rowcolor{DnCBG} JEST++&B&\hspace{0.5em}4B&       233&\hspace{0.5em}23&\textbf{\hspace{0.1em}$+$ 2.8}&\textbf{70.3}&\textbf{76.9}&\textbf{70.3}&\textbf{53.3}\\
\rowcolor{DnCBG} Flexi-JEST++&B&\hspace{0.5em}4B&       110&\hspace{0.5em}\textbf{11}&\hspace{0.1em}$+$ 0.9&68.2&75.8&68.0&51.2\\

% &&&&&&&&&\\[-0.5em]
% \rowcolor{DnCBG} JEST++ &B&10B&       233&58&4.0&72.3&77.6&71.8&53.9\\
% \rowcolor{DnCBG} Flexi-JEST++ &B&10B&       110&28&3.1&71.1&77.2&70.2&53.3\\

&&&&&&&&&\\[-0.5em]
CLIP \cite{radford2021learning}&L&13B&       100 & \hspace{0.5em}32 & -- 11.0 & & 75.5 & 56.3 & 36.5\\
EVA-CLIP \cite{sun2023eva}&L&\hspace{0.5em} 4B &       100 & \hspace{0.5em}10 & \hspace{0.4em}-- 3.4 & & 79.8 & 63.7 & 47.5\\
OpenCLIP \cite{ilharco_gabriel_2021_5143773}&L&32B&       100 & \hspace{0.5em}80 & \hspace{0.4em}-- 6.3 & & 74.0 & 62.1 & 46.1\\
SigLIP \cite{zhai2023sigmoid} &L&40B&       100&100&\hspace{1.1em}0.0&77.1&80.5&69.5&51.2\\
%\rowcolor{DnCBG} JEST++ &L&\hspace{0.5em}4B&       817&82&1.2&75.6&80.1&72.0&55.4\\
\rowcolor{DnCBG}JEST++ &L&\hspace{0.5em}4B&       233 & \hspace{0.5em}\textbf{23} & \textbf{\hspace{0.1em}$+$ 1.8} & 75.5 & \textbf{80.5} & \textbf{71.1} & \textbf{54.8} \\

\end{tabular}
% }
% End paste

%\end{tabular}
% \vspace{1em}
\caption{
\textbf{Comparison to prior art.} FLOP \% are measured relative to SigLIP \cite{zhai2023sigmoid}. Mean denotes the average performance over all metrics. ``Per Iter.'' denotes FLOPs per iteration. 
}
% and colour gradients 
\label{tab:main_table}
\end{table*}

%% file: Tables/laion_table.tex
\noindent
\begin{minipage}{0.53\textwidth}
%\centering
\small
\begin{tabular}{lcccc|c|cccc}
Method & \# Train & IN1K ZS & COCO \\
\hline
% Data goes here
LAION-440M \citep{radenovic2023filtering} & 12.8B & 64.1 & 48.1  \\ 
SemDeDup \citep{abbas2023semdedup} & 8.8B & 64.3 & 48.9 \\
DBP \citep{abbas2024effective} & 5.3B & 65.5 & 48.4 \\
DBP \citep{abbas2024effective} & 3.6B & 64.1 & 45.7 \\
SigLIP \cite{zhai2023sigmoid} & 1.3B & 57.2 & 43.3 \\
\rowcolor{DnCBG}JEST++ & \textbf{1.3B} & \textbf{66.8} & \textbf{54.8} \\
\end{tabular}
\end{minipage}%
\hspace{0em}%
\begin{minipage}{0.47\textwidth}
\vspace{1.5em}
\captionsetup{type=table}
\captionof{table}{\textbf{Comparison to LAION pretraining.} JEST++ strongly surpasses prior art while requiring significantly fewer training iterations. COCO performance denotes the average of image-to-text and text-to-image retrieval.}
\label{tab:test}
\end{minipage}
\vspace{0em}

%% file: Tables/rawvsfiltered.tex
\noindent
\begin{minipage}{0.5\textwidth}

\definecolor{goodgreen}{rgb}{0.25,0.7,0.25}
\definecolor{goodred}{rgb}{0.9,0.25,0.25}

\pgfmathdeclarefunction{cliptoone}{1}{%
  \pgfmathparse{ifthenelse(#1 > 1, 1, ifthenelse(#1 < -1, -1, #1))}%
}

\newcommand{\coloredcell}[3]{%
    \pgfmathsetmacro{\maxval}{#2}%
    \pgfmathsetmacro{\minval}{#3}%
    \pgfmathsetmacro{\scalar}{#1}%
    \pgfmathsetmacro{\range}{\maxval - \minval}%
    \pgfmathsetmacro{\normscalar}{(\scalar - \minval) / (\maxval - \minval) * 2 - 1}%
    \pgfmathsetmacro{\clippednorm}{cliptoone(\normscalar * 0.8)}%
    \pgfmathsetmacro{\opacity}{100 * abs(\clippednorm) * 1.0}%
    \ifdim\normscalar pt > 0pt
        \edef\tempcell{\noexpand\cellcolor{goodgreen!\opacity!white}}%
        \tempcell \scalar%
    \else
        \edef\tempcell{\noexpand\cellcolor{goodred!\opacity!white}}%
        \tempcell \scalar%
    \fi
}
\newcommand{\coloredcellpositive}[3]{%
    \pgfmathsetmacro{\textscalar}{#1}%
    \pgfmathsetmacro{\maxval}{#2}%
    \pgfmathsetmacro{\valscalar}{#3}%
    \pgfmathsetmacro{\normscalar}{\valscalar}%
    \pgfmathsetmacro{\opacity}{100 * log10(\valscalar) * 0.8}%
    \edef\tempcell{\noexpand\cellcolor{goodgreen!\opacity!white}}%
    \tempcell \textscalar%
}

\newcommand{\coloredcellnegative}[3]{%
    \pgfmathsetmacro{\textscalar}{#1}%
    \pgfmathsetmacro{\maxval}{#2}%
    \pgfmathsetmacro{\valscalar}{#3}%
    \pgfmathsetmacro{\normscalar}{\valscalar}%
    \pgfmathsetmacro{\opacity}{100 * log10(\valscalar) * 0.8}%
    \edef\tempcell{\noexpand\cellcolor{goodred!\opacity!white}}%
    \tempcell \textscalar%
}

\small 

\begin{tabular}{lcccc|c|cccc}
Method & Filtered Data? & IN1K ZS & COCO \\
\hline
\small
% Data goes here
\multirow{2}{*}{IID} & \checkmark & 73.6 & 52.5  \\ 
 & \xmark & \coloredcell{69.4}{83.6}{63.6} & \coloredcell{49.5}{62.5}{42.5} \\
\hline
\multirow{2}{*}{JEST++} & \checkmark & \coloredcell{76.9}{83.6}{63.6} & \coloredcell{61.8}{62.5}{42.5} \\
 & \xmark & \coloredcell{76.7}{83.6}{63.6}  & \coloredcell{61.8}{62.5}{42.5}  \\
\end{tabular}
\end{minipage}%
\begin{minipage}{0.5\textwidth}
\vspace{1.0em}
\captionsetup{type=table}
\captionof{table}{\textbf{Simplifying data curation.} All models are trained for 4B examples seen. Performance for JEST++ is nearly identical for pre-training on raw (uncurated) vs. filtered data. Color gradients measured relative to IID training on filtered WebLI.}
\label{tab:rawvsfilt}
\end{minipage}

\vspace{-1em}

%% file: Sections/discussion.tex
\section{Discussion}

We proposed a method---JEST---for jointly selecting the most learnable batches of data, which significantly accelerates large-scale multimodal learning, surpassing the previous state-of-the-art with up to 10$\times$ fewer FLOPs and 13$\times$ fewer examples. In particular, our experiments point to the strong potential for ``data quality bootstrapping'', using small curated datasets to guide learning on much larger, uncurated ones. 
% Our experiments suggest a notion of ``data quality'' that is transferable across scale, represented in this work by small curated reference models that are used to amplify the usefulness of large web-scraped datasets. 

% %While great progress has been made in pre-filtering of large-scale datasets 
% While pre-filtered \textit{foundation datasets} can result in training efficiency gains,
% %into smaller \textit{foundation datasets}, 
% our work adds to a growing body of work \cite{goyal2019recurrent} that highlights
% %that performance improvements result by dynamically composing good batches.
% the limitations of this approach
% when . 
% Rather than \textit{filtering} dataset examples, we instead advocate for scoring the dataset once and sampling batches of examples accordingly. This static \textit{foundation distribution} approach naturally extends to \textit{dynamic curation} which takes into account the state of the learner, as is the case for learnability scoring we explore in this paper. 

Recent work has shown that static dataset filtering, without knowledge of downstream training, can ultimately limit performance \cite{goyal2024scaling}. Our results %further 
demonstrate that 
%dynamically constructing 
useful batches, which must be constructed online, improve pretraining efficiency beyond individually selected examples. These findings therefore advocate for \textit{foundation distributions}---either through pre-scored datasets with \textit{easy-reference} JEST, or dynamically adjusted to the demands of the model with \textit{learnability} JEST---as a more general and effective replacement to generic foundation datasets. 

% pre-scored datasets - 

\textbf{Limitations.} While our method has accelerated multimodal learning of canonical downstream tasks, it has relied on small, well-curated reference datasets which specify the distribution to prioritize within much larger uncurated data. We would therefore encourage future work exploring the inference of reference datasets from the set of downstream tasks of interest.

% We primarily show the impact of our method in the context of foundation models that generalize across multiple downstream evaluations, but Table \ref{tab:main_table} also shows that the choice of reference model 
% (and accordingly the curation datset) 
% can impact specialization to certain downstream evals. 
%For example, JEST variants outperform JEST++ \cut{on the Cars benchmark}, despite under-performing on average. 
% \cut{Further analysis of the performance of reference models and their JEST counterparts in Fig. \ref{fig:fig_appendix_curation}, reveals that these specialization effects can be attributed to the specialization of their reference models to specific domains. These results suggest that our method may enable even larger gains in a transfer-centric approach, specifically choosing reference datasets to target important downstream evaluations such as fairness or under-resourced groups.} 
% In sum, JEST reinforces the potential for active learning to radically transform the efficiency, flexibility, and accessibility of foundation model training.  % Together with JEST's ability to accelerate learning for multimodal generalists, 
% We plan to explore this in future work along with further detailing the scalable properties of ``data quality''. Future work will also investigate whether joint example selection can be applied to accelerate other kinds of learning paradigms beyond contrastive learning.  

% \talfan{Add limitations of method.}

%% file: Sections/appendix.tex
\section{Appendix}

\subsection{Training configuration}

\label{sec:app-config}
Our default training configuration follows that of SigLIP \cite{zhai2023sigmoid}, with a ViT-B/16 and Bert-B image-text dual encoder, training on WebLI for 3 billion examples with a batch size of 32k and the sigmoid-contrastive loss. The vision encoder takes images resized to (256 x 256) and the text-encoder tokenizes text with the sentencepiece tokenizer \cite{kudo2018sentencepiece} trained on the English C4 dataset \cite{raffel2020exploring}. We crop the text to the first 64 tokens. The initial learning rate is 0.001, warmed up linearly during the first 1\% of training, followed by cosine decay. We use a weight decay 0.0001, gradient clipping to a maximum norm of 1.0, and the Adam optimizer with $\beta_2=0.95$. We split training across 256 TPUv5e chips.

For the LAION experiments in Table. \ref{tab:test}, we use an architecture matched to the prior art (ViT-B/32 vision encoder and resized image inputs to (224 x 224)). Otherwise, we use the same training settings as above. We note that this batch size is comparable to that used in the prior art and we find similar results using the softmax-contrastive loss instead of the sigmoid-contrastive loss.

\input{Algorithms/training_loop}

\subsection{Optimisation robustness} \label{sec:beta_opt}

\begin{figure*}[h]
    \centering
    \includegraphics[width=1.0\linewidth]{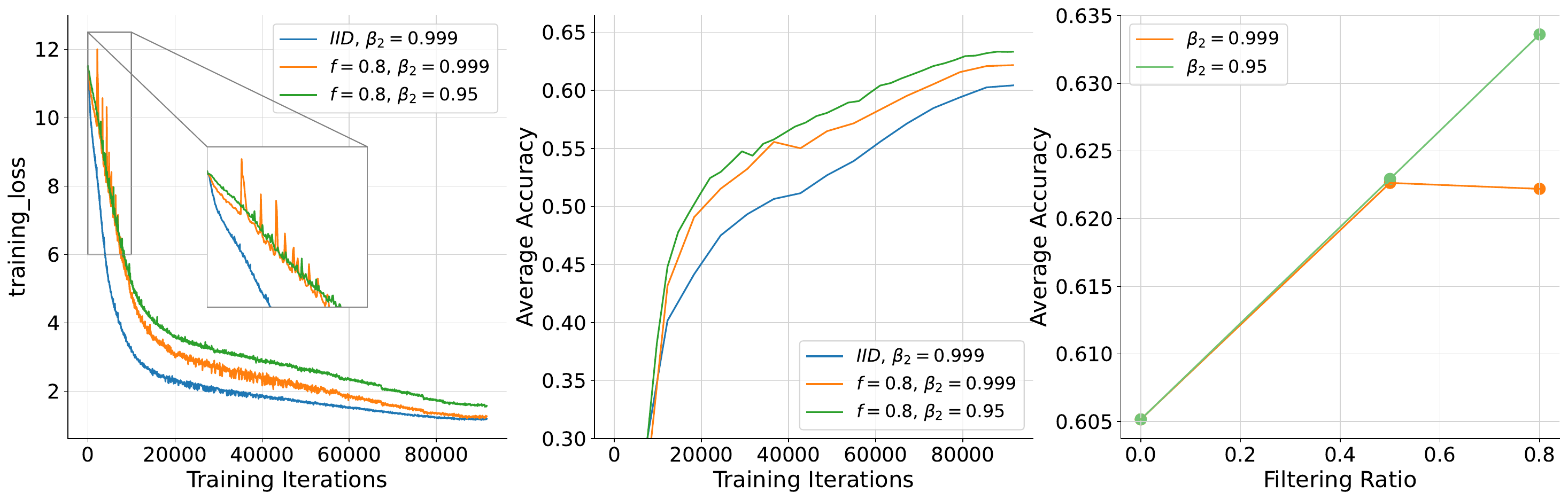}
    \caption{\textbf{Aggressive data prioritization requires robust optimization to perform well. Left:} With standard optimization settings (e.g. $\beta_2 = 0.999$ in the Adam optimizer), aggressive data prioritization leads to instabilities (spikes) in the optimization process. Setting $\beta_2$ to the more stable value of 0.95 remediates this instability. \textbf{Middle, Right: } Stable optimization and strong prioritization together yield large improvement gains.
    }
    \label{fig:fig_optimisation}
\end{figure*}

Our initial experiments used the default ADAM optimizer parameters. At a filtering ratio of 50\%, we observed good gains over IID as described previously. Increasing the filtering ratio to 80\% however did not produce further gains. Inspecting the training curves showed training instabilities (Figure \ref{fig:fig_optimisation}, Left). In the original SigLIP paper \cite{zhai2023sigmoid}, the authors found similar instabilities when increasing the batch size, but found that setting $\beta_2=0.95$ (from $\beta_2=0.999$) ameliorated the problem.

We find the same behaviour (Figure \ref{fig:fig_optimisation}). Although the filtering ratio does not directly affect the training batch size (which is constant in our experiments), this finding suggests that filtering for salient data has the effect of increasing the \textit{effective} batch size. Although this suggests an importance sampling interpretation, applying a re-weighting to the selected data decreased performance in our experiments. We leave this for further work - it is possible that further performance could be extracted from optimizer tuning at higher filtering ratios.

\subsection{Effects of varying the training batch size} \label{sec:effective_batch_size}

It is well known that performance improvements saturate with increased training batch size. In \citep{zhai2023sigmoid}, increasing the batch size beyond 32K was found to actually decrease performance, even after adjusting the $\beta_2$ parameter. In our experiments, we use 32K batch size throughout.

The observation that filtering larger amounts of data produced the same loss spikes as observed by \cite{zhai2023sigmoid} suggests that the training batches selected by JEST might correspond to a much larger \textit{effective} batch size. To investigate, we conducted an ablation in which we instead fixed the super-batch size and progressively decreased the training batch size (i.e. changing the filtering ratio by decreasing the amount of training data, instead of increasing the size of the super-batch as done throughout the paper).

\begin{figure*}[h]
    \centering
    \includegraphics[width=\linewidth]{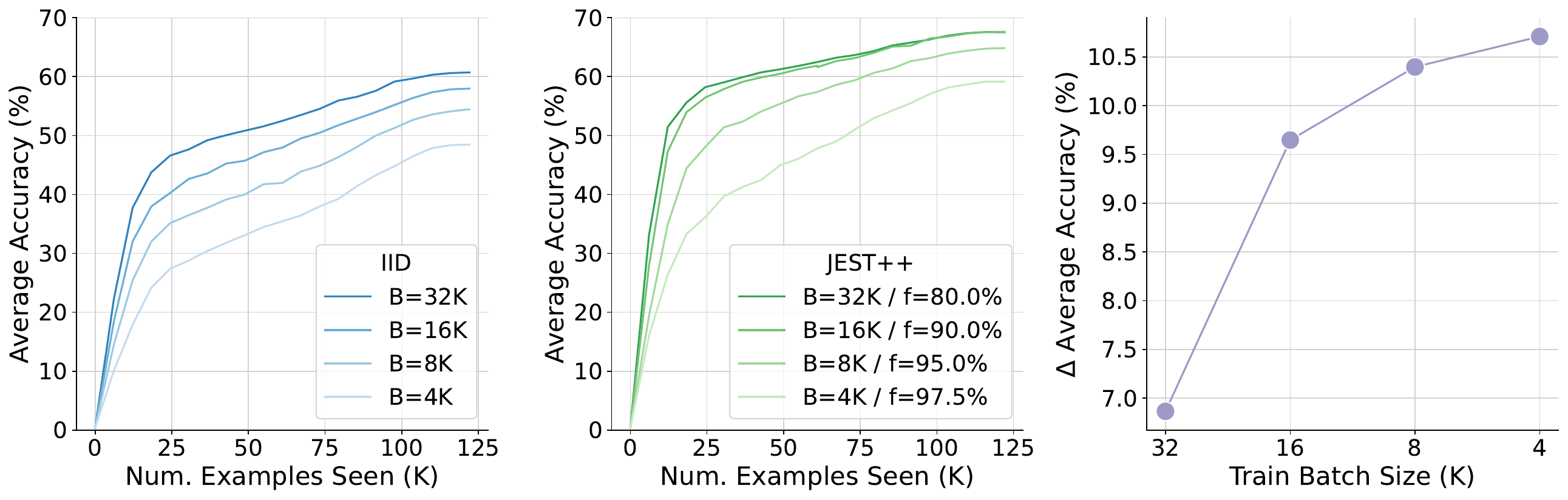}
    \caption{\textbf{Effective Batch Size Experiments} Right: Difference in average performance between JEST++ and IID training as a function of training batch size. Instead of increasing the filtering ratio by increasing the super-batch size as done in previous experiments, we instead fix the super-batch size and reduce the training batch size. There was no noticeable drop from halving the training batch size, suggesting that further efficiency gains might be achieved by training on less data in addition to filtering from a larger pool.}
    \label{fig:fig_effective_batch_size}
\end{figure*}

The results in Figure \ref{fig:fig_effective_batch_size} demonstrate that, as we decrease the batch size (increase the proportion of data filtered) for a fixed super-batch size of 160K, the performance drops predictably for IID training (Left) but decreases much more slowly for JEST training (Middle / Right). Notably, for a halving of the batch size from 32K (corresponding to our f=80\% experiments throughout) to 16K (filtering 90\%), there was no noticeable performance drop.

These results suggest that, at 32K training batch size, our experiments might be already operating at close to the optimal \textit{effective} batch size. We did not conduct further ablations, but it is possible that a more favourable FLOP improvement could be achieved by simultaneously increasing the super-batch size and decreasing the training batch size.

These results suggest an importance sampling interpretation of learnability scoring - assuming the ``True'' mini-batch gradient is given by the expectation of the gradients from IID samples from the data, JEST is sampling only the data that contributes most to that expectation. This suggests that most of the gradient information can be reconstructed from a small number of data points. Although JEST does not explicitly sample based on the magnitude of the gradients of the data, it was demonstrated previously via a simple Taylor expansion argument that the two are equivalent in the case where learnability scores are small \cite{evans2023bad}.

\subsection{FLOP calculations for JEST / Flexi-JEST} \label{sec:flop_calcs}

We assume that training on a single data point cost approximately $C_{\text{IID}}=3F$ forward passes $F$ of the learner model \cite{jouppi2017datacenter}. The cost for a single JEST update can therefore be computed as:

$$
C_{\text{JEST}} = 3F + FB / b - F = F(2 + B/b)
\label{eq:JEST_flops}
$$

where $B$ and $b$ are the super-batch and sub-batch sizes respectively and $f = 1 - b/B$. The base JEST method does not use approximations on the learner, which allows us to cache the forward pass during scoring and re-use it for the gradient computation. Relative to an IID update, the cost of a single JEST iteration at a filtering ratio $f=0.8$ comes out as $\alpha_{\text{JEST}} = 7/3 = 2.33$. Flexi-JEST uses two approximations. Firstly, we split the training batch 50:50 between the full:approximate learner, effectively parallelising the method of \cite{li2023scaling} and reducing the per-data point cost of training. Secondly, we approximate the learner when performing scoring, which reduces the cost of scoring. The overall cost for a single Flexi-JEST update can therefore be computed as:

$$
C_{\text{Flexi-JEST}} = 3F(0.5 + 0.5 A) + AFB/b
$$

where $A$ is the FLOP reduction factor resulting from model approximation (i.e. increasing the patch size, see \ref{sec:approx_multires}). Note that we can no longer cache the forward pass from scoring since it is computed with an approximate version of the learner. Relative to an IID update, the cost of a single Flexi-JEST iteration at a filtering ratio $f=0.8$ and approximation factor $A=0.25$ comes out as $\alpha_{\text{Flexi-JEST}} = 1.04$. In practice, \cite{li2023scaling} estimated the FLOP reduction from a doubling in patch size as closer to $A=0.28$, which is slightly higher than the $A=0.25$ expected by reducing the number of patches by $0.5^2$. We use this more conservative calculation ($\alpha_{\text{Flexi-JEST}} = 1.10$) throughout.

\subsection{Caching reference model scores} \label{sec:ref_cache}

Since the reference model is pretrained and fixed, its scores do not vary over the course of a training run and can be cached within the dataset. For independent example selection we only need to store the scalar scores. However, since data is not likely to be sampled from the training set in the same order in which it is initially scored (e.g. if the batch size varies), the batch composition is unknown ahead of training, which will affect the computation of the scores.

To amortize the cost of reference model scoring across training runs for joint example selection, we therefore instead store the \textit{embeddings} from the reference mode. For a ViT-B/16, the embeddings (text and image) are of size 768, which are considerably smaller but not negligible in comparison to the raw data points. Given these embeddings, the super-batch contrastive matrix can be recomputed before sub-sampling to obtain the sub-batch.

%\input{Algorithms/training_loop}

% def sample_inds(logits):
%   n_images = logits.shape[0]
%   n_draws = int(n_images * samples_per_insert / n_chunks)
%   return random.choice(n_images, n_draws, replace=False, logits=logits)

% def gather(inds, x):
%   inds = np.concatenate(inds, axis=0)
%   return x[inds]

\subsection{Contrastive loss ablations}
\label{sec:softmax}

Contrastive learning maximizes the alignment of image and text modalities for paired examples, while minimizing the alignment of unpaired examples, with batch-level losses $\ell( \mathcal{B} | \theta ) = \frac{1}{b} \sum_{i=1}^b \ell(\vx_i | \theta, \mathcal{B} )$. 

Each data point $\vx_i$ is comprised of an image and associated text which are embedded with their respective encoders as 
$\vz^\textrm{im}_i = f^\textrm{im}(\vx_i ; \theta)$ and $\vz^\textrm{txt}_i = f^\textrm{txt}(\vx_i ; \theta)$. 

In softmax-contrastive learning \cite{radford2021learning}, the conditional loss is 
\begin{equation}
\label{eq:con}
\ell(\vx_i | \theta, \mathcal{B} ) = 
- \frac{1}{2} \left( \log \frac{\exp( \alpha \vz^\textrm{im}_i {\cdot} \vz^\textrm{txt}_i)}{\sum_j \exp( \alpha \vz^\textrm{im}_i {\cdot} \vz^\textrm{txt}_j )}
+ \log \frac{\exp( \alpha \vz^\textrm{im}_i {\cdot} \vz^\textrm{txt}_i)}{\sum_j \exp( \alpha \vz^\textrm{txt}_i {\cdot} \vz^\textrm{im}_j )} \right)
\end{equation}
whereas in sigmoid-contrastive learning \citep{zhai2023sigmoid}, the conditional loss is
\begin{equation}
\label{eq:sig}
\ell(\vx_i | \theta, \mathcal{B} ) =
\log \left[ 1 + \exp( - \alpha \vz^\textrm{im}_i {\cdot} \vz^\textrm{txt}_i + \beta ) \right] + \sum_{j \neq i} \log \left[ 1 + \exp( \alpha \vz^\textrm{im}_i {\cdot} \vz^\textrm{txt}_j - \beta ) \right].
\end{equation}

\begin{figure*}[t]
    \centering
    \includegraphics[width=0.8\linewidth]{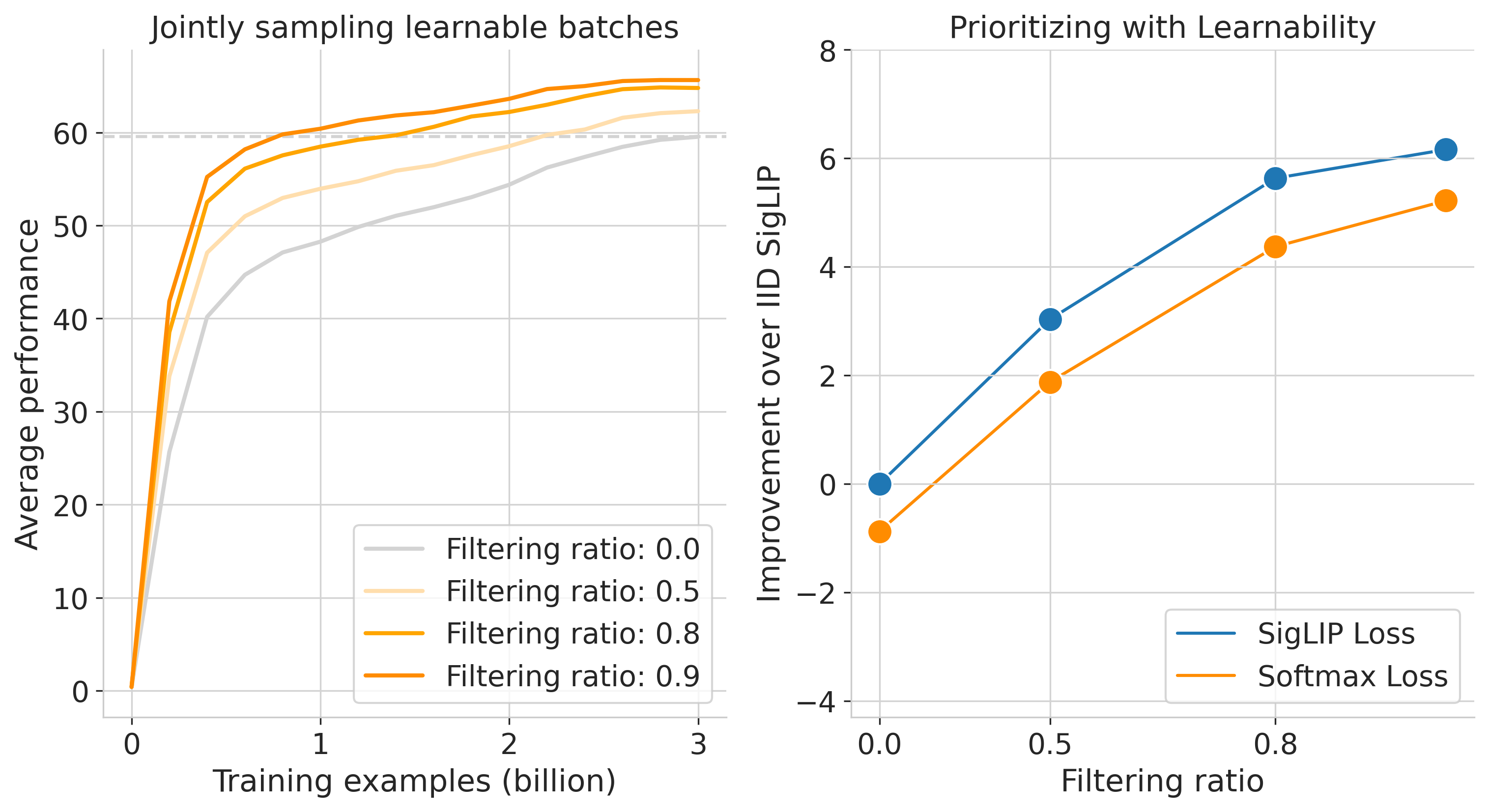}
    \caption{\textbf{JEST is robust to the choice of contrastive loss (softmax vs. sigmoid).} We confirm the robustness of JEST to variations of contrastive learning by testing with the standard softmax-based loss. \textbf{Left:} Similar to the SigLIP version, softmax contrastive learning is accelerated by JEST and benefits from increasing filtering ratios. \textbf{Right:} Compared to the SigLIP loss, the uniform sampling softmax baseline is slightly worse, but relative improvements from JEST are maintained.}
    \label{fig:fig_softmax}
\end{figure*}

Although we leverage the sigmoid pairwise contrastive loss (SigLIP) formulation for our main results, a natural question is whether JEST benefits the standard softmax contrastive learning in Eq. \ref{eq:con}. Due to the formulation of the loss, the JEST algorithm is slightly different from the sigmoid version. We detail the joint-example selection algorithm in Alg. \ref{alg:softmax}, but the main training loop remains the same as in Alg. \ref{alg:training_loop}. We note that unlike the sigmoid version, the softmax loss requires re-computing the softmax during the iterative conditional sampling. This leads to inefficiencies relative to the SigLIP formulation, especially when the batch is split over a large number of devices.

\input{Algorithms/training_loop_softmax}

Nevertheless, in Fig. \ref{fig:fig_softmax}, we show that JEST is indeed robust to the choice of contrastive loss. In the right panel, we see that the gains over the baseline softmax are comparable to the gains for the SigLIP loss. However, due to the degradation in the softmax baseline relative to the baseline SigLIP, the combination of JEST with SigLIP is preferred. 

\subsection{Comparing approximation methods}
\label{sec:app-approximation}

We compared two canonical strategies for online model approximation. Both ablations are conducted at $\sim$75\% FLOP reduction by either dropping 75\% of patches or doubling the patch size (see Main Section \ref{sec:methods}). % (see Main Section \ref{sec:methods})
We vary the proportion of data used for approximate and full-resolution training ($\lambda$ and $1 - \lambda$, respectively), keeping the total number of FLOPs used by the learner the same. Since the cost of one training iteration is proportional $0.25\lambda + 1 - \lambda$, we divide the number of training iterations by this factor to keep the training budget constant with respect to $\lambda$. For $\lambda \in [0.0, 0.25, 0.5, 0.75, 0.95]$, this results in training budgets of $[3, 3.69, 4.8, 6.86, 10.43]$ billion examples seen.

\begin{figure*}[h]
    \centering
    \includegraphics[width=0.85\linewidth]{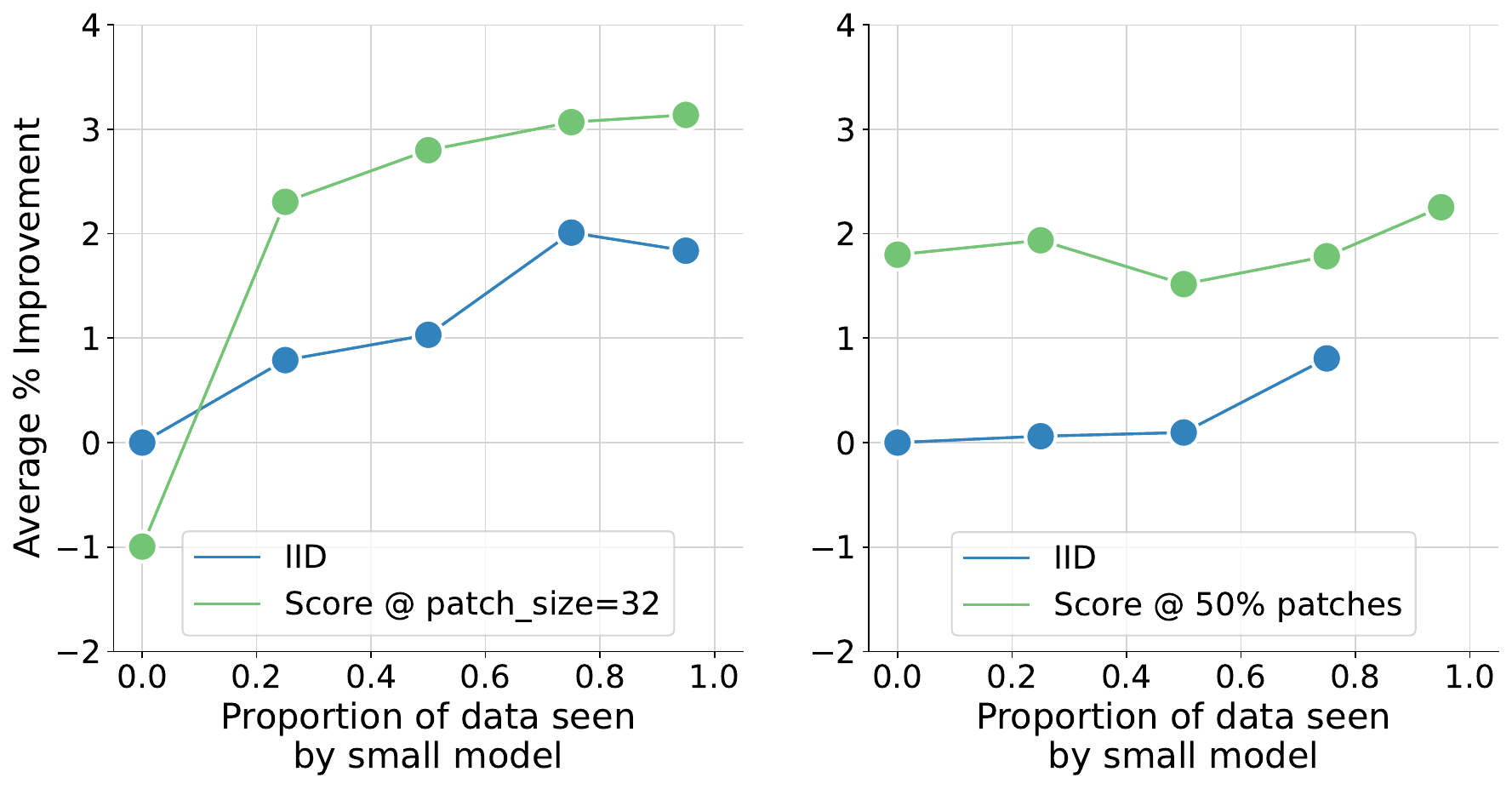}
    \caption{\textbf{Comparing model approximation strategies.} Comparing Flexi-JEST and PatchDrop-JEST at a 50\% filtering proportion. On the x-axis, we vary the proportion of data used to co-train the scoring model. Green curve shows JEST runs, blue curves shows equivalent IID run without data curation. All runs conducted at isoFLOP, see Fig. \ref{fig:fig_multires}. \textbf{Left: } FlexiViT scoring performs badly 0-shot out of distribution inference, but quickly recovers with co-training. \textbf{Right: } Patch dropping is more robust to 0-shot inference, but doesn't benefit as much from co-training.}
    \label{fig:fig_appendix_multires}
\end{figure*}

Our results (Figure \ref{fig:fig_appendix_multires}) demonstrate that downscaling data by decreasing the resolution effects a much more favourable trade-off than dropping a subset of patches. Both methods perform differently out-of-distribution, but only FlexiViT benefits significantly from co-training. We adopt this strategy throughout the paper.

% Curation study figure
% \begin{figure*}[h]
%     \centering
%     \includegraphics[width=0.5\linewidth]{Figures/fig_appendix_curation_study.png}
%     \caption{\textbf{Data curation can steer reference models to perform well in specific domains, which is correlated with the specialized performance of corresponding JEST models.} We unpack the averaged results in Fig. \ref{fig:fig_curation} (left) by displaying performance across 4 evals (ImageNet-ZS, ImageNet-10-shot, COCO I2T, and COCO T2I). Weakly curated datasets do not show any correlation across evals due to the lack of any strong effect on active learning. However, with moderate curation, there is a strong correlation between reference model and JEST performance across evaluations. This indicates that depending on the type of curation, one can specialize references to certain domains which then \textit{transfers specialization to the learner model}. The correlation is also visible for strongly curated data; however, this type of curation seems to generalize better across domains.}
%     \label{fig:fig_appendix_curation}
% \end{figure*}

% TIGG Hard Learner
\begin{figure*}[h]
    \centering
    \includegraphics[width=0.5\linewidth]{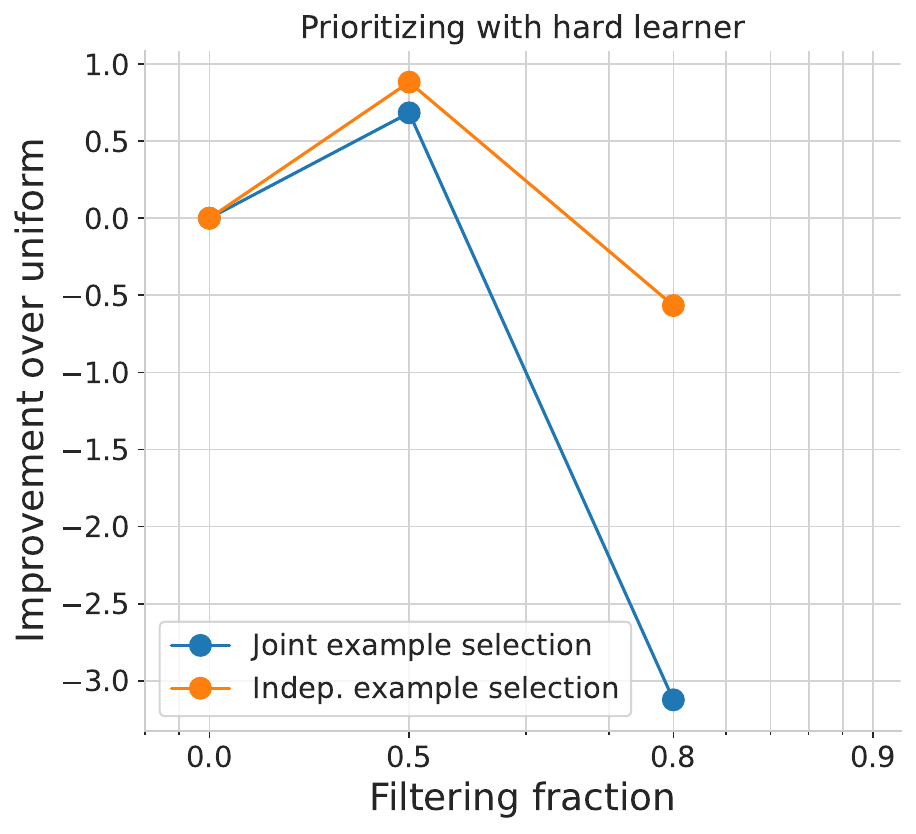}
    \caption{\textbf{Prioritizing data that is difficult for the learner} Compared to \textit{easy-reference} and \textit{learnability} scoring (Figure \ref{fig:fig_pairwise}) prioritising data with high learner loss results in small gains at a filtering ratio of 50\%, but quickly degrades as we filter larger amounts of data. Joint example selection exacerbates the effect at larger filtering ratios, aligning with the interpretation that \textit{hard-learner} prioritisation prioritises sampling noise in the data.
    }
    \label{fig:tigg_hard_learner}
\end{figure*}

% Expanded results Table
\input{Tables/appendix_table}

\input{Tables/appendix_table_2}

%% file: Algorithms/training_loop.tex
\begin{algorithm}[t]
\caption{Pseudocode for JEST / Flexi-JEST}
\label{alg:training_loop}
\definecolor{codeblue}{rgb}{0.25,0.5,0.5}
\definecolor{codegreen}{rgb}{0.25,0.5,0.25}
\lstset{
  backgroundcolor=\color{ouralg!20},
  basicstyle=\fontsize{7.2pt}{7.2pt}\ttfamily\selectfont,
  columns=fullflexible,
  breaklines=true,
  captionpos=b,
  commentstyle=\fontsize{7.2pt}{7.2pt}\color{codeblue},
  keywordstyle=\fontsize{7.2pt}{7.2pt}\color{codeblue},  % \color{codeblue}
}
\begin{lstlisting}[language=python]

cfg = ConfigDict(
    n_chunks=16,
    filter_ratio=0.8,
    method="learnability",
    method="jest",  # or "flexi-jest"
    softmax_score_gain=100.0,
    loss_type = "sigmoid",
)

def sigmoid_nll(params, embeds):
  zimg, ztxt = embeds
  logits = np.dot(zimg, ztxt.T)  # [B, B]
  logits = logits * params["alpha"] + params["beta"]
  eye = np.eye(zimg.shape[0])
  m1_diag1 = -np.ones_like(logits) + 2 * eye
  
  nll_mat = -log_sigmoid(m1_diag1 * logits)
  nll = np.sum(nll_mat, axis=-1).mean()

  return nll, nll_mat  # [,], [B, B]

def get_scores_sigmoid(embeds, embeds_ref, params, params_ref):
  _, nll_mod = sigmoid_nll(params, embeds)  # [B, B]
  _, nll_ref = sigmoid_nll(params_ref, embeds_ref)  # [B, B]
  if cfg.scoring == "learnability":
    scores = nll_mod - nll_ref
  elif cfg.scoring == "easy_ref":
    scores = - nll_ref
  return scores * cfg.softmax_score_gain
 
def loss_fn(params, params_ref, batch):
  images, texts = batch
  approx = True if cfg.method == "flexi-jest" else False
  
  # Score and sub-sample the initial super-batch
  embeds = model.forward(images, texts, params, approx=approx)  # [5B, D]
  embeds_ref = batch["embeds_ref"]  # Pre-cached in dataset
  if cfg.loss_type == "sigmoid":
      scores = get_scores_sigmoid(embeds, embeds_ref, params, params_ref)  # Get scores
      inds = jointly_sample_batch(scores, cfg.n_chunks, cfg.filter_ratio, cfg.learnability)
  elif cfg.loss_type == "softmax":
      inds = jointly_sample_batch_softmax(embeds_ref, embeds, params_ref, params, cfg.n_chunks, cfg.filter_ratio) # for softmax loss, scores are re-computed in the iterative sampling. 

  images, texts = stop_grad(images[inds]), stop_grad(texts[inds])  # [B, ...]
  
  # Split batch for co-training
  images_full, images_approx = images[::2], images[1::2]  # [B/2, ...], [B/2, ...]
  texts_full, texts_approx = texts[::2], texts[1::2]  # [B/2, ...], [B/2, ...]
  
  # Compute overall loss
  embeds_full = model.forward(images_full, texts_full, params, approx=False)  # [B/2, D], [B/2, D]
  embeds_approx = model.forward(images_approx, texts_approx, params, approx=approx)  # [B/2, D], [B/2, D]
  zimg = np.concatenate([embeds_full[0], embeds_approx[0]], axis=0)
  ztxt = np.concatenate([embeds_full[1], embeds_approx[1]], axis=0)
  if loss_type == "sigmoid":
      loss, _ = sigmoid_nll(params, (zimg, ztxt))
  elif loss_type == "softmax":
      loss, _, _ = softmax_nll(params, (zimg, ztxt), is_sampled=None)
 
  return loss

\end{lstlisting}
% \vspace{1em}
\end{algorithm}

%% file: Algorithms/training_loop_softmax.tex
\begin{algorithm}[t]
\caption{Joint example selection: softmax loss}
\label{alg:softmax}
\definecolor{codeblue}{rgb}{0.25,0.5,0.5}
\definecolor{codegreen}{rgb}{0.25,0.5,0.25}
\lstset{
  backgroundcolor=\color{ouralg!20},
  basicstyle=\fontsize{7.2pt}{7.2pt}\ttfamily\selectfont,
  columns=fullflexible,
  breaklines=true,
  captionpos=b,
  commentstyle=\fontsize{7.2pt}{7.2pt}\color{codeblue},
  keywordstyle=\fontsize{7.2pt}{7.2pt}\color{codeblue},  % \color{codeblue}
}

\lstset{keepspaces=true}
\begin{lstlisting}[language=python]
def softmax_neg(logits, is_sampled=None, axis=-1):
    # during the iterative sampling, we need to mask out negatives that have already been selected
    if is_sampled is not None:
        logits_neg = logits - (1.- is_sampled)*1e8
    else:
        logits_neg = logits
    logits_neg = nn.logsumexp(logits_neg, axis=axis) # Compute the softmax negative term
    return logits_neg

def softmax_nll(params, embeds, is_sampled=None):
    zimg, ztxt = embeds
    logits_mat = np.dot(zimg, ztxt.T) * params["t"]
    if is_sampled is not None:
        is_sampled_0 = is_sampled.reshape(logits_mat.shape[0], 1)
        is_sampled_1 = is_sampled.reshape(1, logits_mat.shape[0])
    else:
        is_sampled_0 = None
        is_sampled_1 = None
        
    logits_ij = softmax_neg(logits_mat, is_sampled_0, axis=0) # calculate negative softmax term for image to text half of loss
    logits_ji = softmax_neg(logits_mat, is_sampled_1, axis=1) # calculate negative softmax term for text to image half of loss
    loss_0 = -(jnp.diag(logits_mat) - logits_ij)
    loss_1 = -(jnp.diag(logits_mat) - logits_ji)
    neg_logits = 0.5*(logits_ij + logits_ji)

    l = jnp.mean(0.5*(loss_0 + loss_1))
    return l, neg_logits, -logits_mat

def jointly_sample_batch_softmax(embeds_ref, embeds_learner, params_ref, params_learner, n_chunks=16, filter_ratio=0.8):
    _, _, learner_logits = softmax_nll(params_learner, embeds_learner, is_sampled=None)
    _, _, ref_logits = sofmtax_nll(params_learner, embeds_learner, is_sampled=None)
    scores = (learner_logits - ref_logits)*cfg.softmax_score_gain
    n_images = scores.shape[0]                               # scores.shape = [B, B]
    n_draws = int(n_images * (1 - filter_ratio) / n_chunks)  # Size of each chunk.
    logits_ii = np.diag(scores)                              # Self-similarity scores.
    inds = random.choice(logits_ii, n_draws)
    # Sample first chunk.
    
    for _ in range(n_chunks - 1):
        is_sampled = np.eye(n_images)[inds].sum(axis=0) # Binary indicator of current samples [n_images,].
        _, learner_logits_n, _ = softmax_nll(params_learner, embeds_learner, is_sampled=is_sampled)
        _, ref_logits_n, _ = softmax_nll(params_ref, embeds_ref, is_sampled=is_sampled)
        rho_scores_n = (learner_logits_n - ref_logits_n)*cfg.softmax_score_gain
        logits = logits_ii + rho_scores_n      # Conditional learnability given past samples.
        logits = logits - is_sampled * 1e8              # Avoid sampling with replacement.
        new_inds = random.choice(n_images, n_draws, p=np.exp(logits))
        inds = np.concatenate((inds, new_inds))         # Expand the array of indices sampled.
        return inds                                       # Gather and return subset indices.

\end{lstlisting}
\end{algorithm}

%% file: Tables/appendix_table.tex
\def\tbd{\textcolor{red}{TBD}}

% flops_per_image_siglip_B_16 = 35.1 * 3
% flops_per_image_jest_B_16 = 35.1 * (2 + 1 / SPI) if we cache the inference pass
%
% ratio = 7/3 = 2.333 for SPI = 0.2
%
% (3 * (0.5 * 1.0 + 0.5 * 0.25)) + 1 / SPI * 0.25
% = 3 * 0.625 + 0.25 * 5
% = 3.125
%
% = 1.042, so 4.2% overhead

\begin{table*}[h]
\centering
%\footnotesize
% \scriptsize
%\footscriptsize
\small

\newcommand*\rot{\rotatebox{90}}

% Paste output from Colab:
%   https://colab.corp.google.com/drive/1jYI-LEGIaf0gL5RTHff0IQwYVeu_m9MW?usp=sharing

% \definecolor{goodgreen}{rgb}{0.25,0.7,0.25}
% \definecolor{goodred}{rgb}{0.9,0.25,0.25}
\definecolor{goodgreen}{rgb}{1.0,1.0,1.0}
\definecolor{goodred}{rgb}{1.0,1.0,1.0}

\newcommand{\coloredcell}[3]{%
    \pgfmathsetmacro{\maxval}{#2}%
    \pgfmathsetmacro{\minval}{#3}%
    \pgfmathsetmacro{\scalar}{#1}%
    \pgfmathsetmacro{\range}{\maxval - \minval}%
    \pgfmathsetmacro{\normscalar}{(\scalar - \minval) / (\maxval - \minval) * 2 - 1}%
    \pgfmathsetmacro{\opacity}{100 * abs(\normscalar) * 0.8}%
    \ifdim\normscalar pt > 0pt
        \edef\tempcell{\noexpand\cellcolor{goodgreen!\opacity!white}}%
        \tempcell \scalar%
    \else
        \edef\tempcell{\noexpand\cellcolor{goodred!\opacity!white}}%
        \tempcell \scalar%
    \fi
}

\newcommand{\coloredcellpositive}[3]{%
    \pgfmathsetmacro{\textscalar}{#1}%
    \pgfmathsetmacro{\maxval}{#2}%
    \pgfmathsetmacro{\valscalar}{#3}%
    \pgfmathsetmacro{\normscalar}{\valscalar}%
    \pgfmathsetmacro{\opacity}{100 * log10(\valscalar) * 0.8}%
    \edef\tempcell{\noexpand\cellcolor{goodgreen!\opacity!white}}%
    \tempcell \textscalar%
}

\newcommand{\coloredcellnegative}[3]{%
    \pgfmathsetmacro{\textscalar}{#1}%
    \pgfmathsetmacro{\maxval}{#2}%
    \pgfmathsetmacro{\valscalar}{#3}%
    \pgfmathsetmacro{\normscalar}{\valscalar}%
    \pgfmathsetmacro{\opacity}{100 * log10(\valscalar) * 0.8}%
    \edef\tempcell{\noexpand\cellcolor{goodred!\opacity!white}}%
    \tempcell \textscalar%
}

% Paste here

% \resizebox{\linewidth}{!}{

\begin{tabular}{lcccc|ccccc}
%\begin{tabular}{@{\extracolsep{1pt}}lccccccccccc|c@{}}

& & & \multicolumn{2}{c}{FLOPs \%} & 
%\textbf{Mean} 
& \multicolumn{2}{c}{ImageNet-1k} & \multicolumn{2}{c}{COCO} \\ % First row
\cmidrule(lr){4-5} \cmidrule(lr){7-8} \cmidrule(lr){9-10}
% \cline{4-5} \cline{7-8} \cline{9-10}
 \\ [-3.0ex]
%\noalign{\smallskip}
Method & Variant & \# Train & Per Iter. & \textbf{Total} & \textbf{Mean} $\Delta$ & 10-S & ZS &  I2T & T2I % & Birds & Caltech &  Cars & Pets \\% & $\mathbf{\Delta}$ \\  % Second row
\\

% Method&\rot{FLOPs (\% Per Iter.)}&\rot{FLOPs (\% Total)}&\rot{Train Ex. (B)}&\rot{IN 10-Shot}&\rot{IN-1K}&\rot{COCO I2T R1}&\rot{COCO T2I R1}&\rot{Birds 10-Shot}&\rot{Caltech 10-Shot}&\rot{Cars 10-Shot}&\rot{Pets 10-Shot}&\rot{Mean}&$\Delta$ (\%)\\

\hline
% \cmidrule(lr){1-10}

%CLIP \cite{radford2021learning}&B&13B&       100&\hspace{0.5em}32&-- 11.8&&68.3&52.4&33.1\\
%EVA-CLIP \cite{sun2023eva}&B&\hspace{0.5em}8B&       100&\hspace{0.5em}20&\hspace{0.4em}-- 4.6&&74.7&58.7&42.2\\
%OpenCLIP \cite{ilharco_gabriel_2021_5143773}&B&34B&       100&\hspace{0.5em}85&\hspace{0.4em}-- 5.8&&70.2&59.4&42.3\\
%LessIsMore \cite{cao2023less}&B&11B&       100&\hspace{0.5em}28&\hspace{0.4em}-- 5.9&&70.8&58.3&42.5\\
%SILC-S \cite{naeem2023silc}&B&20B&       380&190&\hspace{0.1em}$+$ 0.2&68.9&76.6&66.2&48.7\\
% &&&&&&&&&\\[-0.5em]
SigLIP \cite{zhai2023sigmoid}&B&40B&       100&100&\hspace{1.1em}0.0&70.3&76.7&65.2&47.4\\

\rowcolor{DnCBG} JEST&B&\hspace{0.5em}5B&       233&29&\hspace{0.3em}-- 0.5&68.2&75.5&66.1&47.9\\
\rowcolor{DnCBG} Flexi-JEST&B&13B&       110&36&\hspace{0.1em}$+$ 1.2&69.4&76.6&68.2&50.2\\
\rowcolor{DnCBG} JEST++&B&\hspace{0.5em}4B&       233&\hspace{0.5em}23&\hspace{0.1em}$+$ 2.8&70.3&76.9&70.3&53.3\\
\rowcolor{DnCBG} Flexi-JEST++&B&\hspace{0.5em}4B&       110&\hspace{0.5em}\textbf{11}&\hspace{0.1em}$+$ 0.9&68.2&75.8&68.0&51.2\\

% &&&&&&&&&\\[-0.5em]
\rowcolor{DnCBG} JEST++ &B&10B&       233&58&\textbf{\hspace{0.1em}$+$ 4.0}&\textbf{72.3}&\textbf{77.6}&\textbf{71.8}&\textbf{53.9}\\
\rowcolor{DnCBG} Flexi-JEST++ &B&10B&       110&28&\hspace{0.1em}$+$ 3.1&71.1&77.2&70.2&53.3\\

&&&&&&&&&\\[-0.5em]
%CLIP \cite{radford2021learning}&L&13B&       100 & \hspace{0.5em}32 & -- 11.0 & & 75.5 & 56.3 & 36.5\\
%EVA-CLIP \cite{sun2023eva}&L&\hspace{0.5em} 4B &       100 & \hspace{0.5em}10 & \hspace{0.4em}-- 3.4 & & 79.8 & 63.7 & 47.5\\
%OpenCLIP \cite{ilharco_gabriel_2021_5143773}&L&32B&       100 & \hspace{0.5em}80 & \hspace{0.4em}-- 6.3 & & 74.0 & 62.1 & 46.1\\
SigLIP \cite{zhai2023sigmoid} &L&40B&       100&100&\hspace{1.1em}0.0&77.1&80.5&69.5&51.2\\
%\rowcolor{DnCBG} JEST++ &L&\hspace{0.5em}4B&       817&82&1.2&75.6&80.1&72.0&55.4\\
\rowcolor{DnCBG}JEST++ &L&\hspace{0.5em}4B&       233 & \hspace{0.5em}\textbf{23} & \textbf{\hspace{0.1em}$+$ 1.8} & 75.5 & 80.4 & \textbf{71.1} & \textbf{54.8} \\

\end{tabular}
% }
% End paste

%\end{tabular}
\vspace{1em}
\caption{
\textbf{JEST continues to improve with longer training runs.} FLOP \% is measured relative to SigLIP \cite{zhai2023sigmoid}. Mean denotes the average performance over all metrics. ``Per Iter.'' denotes FLOPs per iteration. 10B training runs of both JEST++ and FlexiJEST++ continue to improve over the 4B results presented in main Table \ref{tab:main_table}. JEST and Flexi-JEST, which use the WebLi-curated reference dataset both perform strongly on a per-FLOP basis, with Flexi-JEST also outperforming the SigLIP 40B baseilne on average.
}

\label{tab:appendix_table}
\end{table*}

%% file: Tables/appendix_table_2.tex
\def\tbd{\textcolor{red}{TBD}}

% flops_per_image_siglip_B_16 = 35.1 * 3
% flops_per_image_jest_B_16 = 35.1 * (2 + 1 / SPI) if we cache the inference pass
%
% ratio = 7/3 = 2.333 for SPI = 0.2
%
% (3 * (0.5 * 1.0 + 0.5 * 0.25)) + 1 / SPI * 0.25
% = 3 * 0.625 + 0.25 * 5
% = 3.125
%
% = 1.042, so 4.2% overhead

\begin{table*}[h]
\centering
%\footnotesize
% \scriptsize
%\footscriptsize
\small

\newcommand*\rot{\rotatebox{90}}

% Paste output from Colab:
%   https://colab.corp.google.com/drive/1jYI-LEGIaf0gL5RTHff0IQwYVeu_m9MW?usp=sharing

\definecolor{goodgreen}{rgb}{0.25,0.7,0.25}
\definecolor{goodred}{rgb}{0.9,0.25,0.25}
%\definecolor{goodgreen}{rgb}{1.0,1.0,1.0}
%\definecolor{goodred}{rgb}{1.0,1.0,1.0}

% Define the cliptoone macro using \pgfmath
\pgfmathdeclarefunction{cliptoone}{1}{%
  \pgfmathparse{ifthenelse(#1 > 1, 1, ifthenelse(#1 < -1, -1, #1))}%
}

\newcommand{\coloredcell}[3]{%
    \pgfmathsetmacro{\maxval}{#2}%
    \pgfmathsetmacro{\minval}{#3}%
    \pgfmathsetmacro{\scalar}{#1}%
    \pgfmathsetmacro{\range}{\maxval - \minval}%
    \pgfmathsetmacro{\normscalar}{(\scalar - \minval) / (\maxval - \minval) * 2 - 1}%
    \pgfmathsetmacro{\clippednorm}{cliptoone(\normscalar * 0.8)}%
    \pgfmathsetmacro{\opacity}{100 * abs(\clippednorm) * 1.0}%
    \ifdim\normscalar pt > 0pt
        \edef\tempcell{\noexpand\cellcolor{goodgreen!\opacity!white}}%
        \tempcell \scalar%
    \else
        \edef\tempcell{\noexpand\cellcolor{goodred!\opacity!white}}%
        \tempcell \scalar%
    \fi
}

\newcommand{\coloredcellpositive}[3]{%
    \pgfmathsetmacro{\textscalar}{#1}%
    \pgfmathsetmacro{\maxval}{#2}%
    \pgfmathsetmacro{\valscalar}{#3}%
    \pgfmathsetmacro{\normscalar}{\valscalar}%
    \pgfmathsetmacro{\opacity}{100 * log10(\valscalar) * 0.8}%
    \edef\tempcell{\noexpand\cellcolor{goodgreen!\opacity!white}}%
    \tempcell \textscalar%
}

\newcommand{\coloredcellnegative}[3]{%
    \pgfmathsetmacro{\textscalar}{#1}%
    \pgfmathsetmacro{\maxval}{#2}%
    \pgfmathsetmacro{\valscalar}{#3}%
    \pgfmathsetmacro{\normscalar}{\valscalar}%
    \pgfmathsetmacro{\opacity}{100 * log10(\valscalar) * 0.8}%
    \edef\tempcell{\noexpand\cellcolor{goodred!\opacity!white}}%
    \tempcell \textscalar%
}

% Paste here

\resizebox{\linewidth}{!}{

\begin{tabular}{lc|cccccccc|c}
%\begin{tabular}{@{\extracolsep{1pt}}lccccccccccc|c@{}}

& & \multicolumn{2}{c}{ImageNet-1k} & \multicolumn{2}{c}{COCO} & & & & & \\  % First row
%\cmidrule(lr){2-3} \cmidrule(lr){5-6} \cmidrule(lr){7-8}
\cmidrule(lr){3-4} \cmidrule(lr){5-6}
 \\ [-3.0ex]
%\noalign{\smallskip}
Method & \# Train & 10-S & ZS &  I2T & T2I & Birds & Caltech &  Cars & Pets & \textbf{Mean} \\% & $\mathbf{\Delta}$ \\  % Second row

% Method&\rot{FLOPs (\% Per Iter.)}&\rot{FLOPs (\% Total)}&\rot{Train Ex. (B)}&\rot{IN 10-Shot}&\rot{IN-1K}&\rot{COCO I2T R1}&\rot{COCO T2I R1}&\rot{Birds 10-Shot}&\rot{Caltech 10-Shot}&\rot{Cars 10-Shot}&\rot{Pets 10-Shot}&\rot{Mean}&$\Delta$ (\%)\\

\hline

\text{SigLIP \cite{zhai2023sigmoid}}& 40B&\textbf{\coloredcell{70.3}{75.3}{65.3}}&\coloredcell{76.7}{81.7}{71.7}&\coloredcell{65.2}{70.2}{60.2}&\coloredcell{47.4}{52.4}{42.4}&\coloredcell{75.5}{80.5}{70.5}&\coloredcell{92.8}{97.8}{87.8}&\coloredcell{92.0}{97.0}{87.0}&\coloredcell{91.2}{96.2}{86.2}&\coloredcell{76.4}{81.4}{71.4}\\
\text{WebLI Curated++ Ref.}&\hspace{0.5em}5B&\coloredcell{60.6}{75.3}{65.3}&\coloredcell{67.1}{81.7}{71.7}&\textbf{\coloredcell{74.2}{70.2}{60.2}}&\textbf{\coloredcell{56.8}{52.4}{42.4}}&\coloredcell{65.4}{80.5}{70.5}&\coloredcell{89.9}{97.8}{87.8}&\coloredcell{64.1}{97.0}{87.0}&\coloredcell{83.0}{96.2}{86.2}&\coloredcell{70.2}{81.4}{71.4}\\
\text{JEST++}&\hspace{0.5em}4B&\textbf{\coloredcell{70.3}{75.3}{65.3}}&\textbf{\coloredcell{76.9}{81.7}{71.7}}&\coloredcell{70.3}{70.2}{60.2}&\coloredcell{53.3}{52.4}{42.4}&\textbf{\coloredcell{80.4}{80.5}{70.5}}&\coloredcell{91.6}{97.8}{87.8}&\coloredcell{89.5}{97.0}{87.0}&\textbf{\coloredcell{93.1}{96.2}{86.2}}&\textbf{\coloredcell{78.2}{81.4}{71.4}}\\

\end{tabular}
}
% End paste

%\end{tabular}
\vspace{1em}
\caption{\textbf{JEST++ efficiently leverages specialist reference models to train generalist foundation models.} Colour gradients are measured relative to SigLIP \cite{zhai2023sigmoid}. Mean denotes the average performance over all metrics. WebLI-curated++ reference performance is highly specialized- improving (over the SigLIP baseline) on only 2 out of the 8 benchmarks and severely under-performing on the rest. On the other hand JEST++ sees far more general improvements across benchmarks (8\% better than the reference model on average). 
}

\label{tab:appendix_table_2}
\end{table*}